\pdfoutput=1
\documentclass[11pt]{article}

\usepackage[final]{naacl2021}

\usepackage{times}
\usepackage{latexsym}

\usepackage[T1]{fontenc}
\usepackage[utf8]{inputenc}

\usepackage{microtype}

\usepackage{algpseudocode}
\usepackage{float}
\usepackage{mathtools}
\usepackage{amsmath}
\usepackage{xcolor}
\usepackage{enumitem}
\usepackage{caption}
\usepackage{subcaption}
\usepackage{wrapfig}
\usepackage{fancybox}
\usepackage{enumitem}
\setlist[itemize]{leftmargin=*,label=\scalebox{.8}{\textbullet}}

\usepackage{url}
\usepackage{booktabs}
\usepackage{hyperref}
\hypersetup{
    colorlinks=true,
    citecolor=black,
    urlcolor=magenta,
}
\usepackage{lineno}
\usepackage{bbm}
\usepackage{arydshln}
\usepackage{fdsymbol}
\usepackage{comment}
\usepackage{pdfpages}
\usepackage{xspace}

\newcommand{\data}{{\sc Mechanic}\xspace}
\newcommand{\kb}{{\sc Comb}\xspace}
\newcommand{\dataE}{{\sc Mechanic-G}\xspace}
\newcommand{\com}[1]{}

\title{Extracting a Knowledge Base of Mechanisms from COVID-19 Papers}

\usepackage{soul}

\DeclareRobustCommand{\gspan}[1]{{\begingroup{\small \textbf{#1}}\endgroup} }

\makeatletter
\newcommand{\printfnsymbol}[1]{%
  \textsuperscript{\@fnsymbol{#1}}%
}
\makeatother

\author{
Tom Hope$^{\clubsuit,\spadesuit}$\thanks{\printfnsymbol{1}Equal contribution.} 
~~~Aida Amini $^\clubsuit$\printfnsymbol{1}
~~~David Wadden$^\clubsuit$ 
~~Madeleine van Zuylen$^\spadesuit$\\\bf
~~Sravanthi Parasa$^\heartsuit$
~~Eric Horvitz$^\diamondsuit$
~~Daniel Weld$^{\clubsuit,\spadesuit}$

~~Roy Schwartz$^{\varheartsuit}$ 
~~Hannaneh Hajishirzi$^{\clubsuit,\spadesuit}$
 \\
$^\clubsuit$ Paul G. Allen School for Computer Science \& Engineering, University of Washington\\
$^\spadesuit$ Allen Institute for Artificial Intelligence ~ 
$^\heartsuit$ Swedish Medical Group \\
$^\diamondsuit$ Microsoft Research ~
$^\varheartsuit$ The Hebrew University of Jerusalem, Israel\\
{\tt \{tomh,aidaa,danw, hannah\}@allenai.org} 
}

\begin{document}
\setcounter{secnumdepth}{3}
%\linenumbers
\maketitle

\begin{abstract}
The COVID-19 pandemic has spawned a diverse body of scientific literature that is challenging to navigate, stimulating interest in automated tools to help find useful knowledge. We pursue the construction of a knowledge base (KB) of \emph{mechanisms}---a fundamental concept across the sciences, which encompasses activities, functions and causal relations, ranging from cellular processes to economic impacts. We extract this information from the natural language of scientific papers by developing a broad, unified schema that strikes a balance between relevance and breadth. We annotate a dataset of mechanisms with our schema and train a model to extract mechanism relations from papers. %dataset of annotated mechanims in scientific papers and train an information extraction model extract mechanisms  a KB of mechanism relations. %\footnote{Data and models are made available at \url{anonymous}.} 
Our experiments demonstrate the utility of our KB in supporting interdisciplinary scientific search over COVID-19 literature, outperforming the prominent PubMed search in a study with clinical experts. Our search engine, dataset and code are publicly available.\footnote{\href{https://covidmechanisms.apps.allenai.org/}{https://covidmechanisms.apps.allenai.org/}}
\end{abstract}

%Proposed revised abstract % dan - waiting til the file calms down
%\begin{abstract}
%The COVID-19 pandemic has spawned a diverse body of scientific literature that is challenging to navigate, stimulating interest in automated tools to help find useful knowledge. We pursue the construction of a knowledge base (KB) of \emph{mechanisms} -- a fundamental concept across the sciences encompassing activities, functions and causal relations, ranging from cellular processes to economic impacts. 

%We annotate a training set of mechanisms and use information extraction to then extract by collecting a dataset of annotated mechanisms and training a model for extracting mechani.  dataset of annotated mechanims in scientific papers and train an information extraction model extract mechanisms  a KB of mechanism relations.

%We extract this information from the natural language of scientific papers, with our developed broad, unified schema  that strikes a balance between expressivity and breadth and enables our KB to generalize across diverse mechanisms. 

 %\footnote{Data and models are made available at \url{anonymous}.} 
%Human evaluation studies show our system retrieves relevant information on viral mechanisms of action and applications of AI against COVID-19 with high precision.
%\end{abstract}
\section{Introduction}
\label{sec:intro}
 \begin{quotation}
\noindent ``Some experts are familiar with one field, such as AI or nanotechnology [...] no one is capable of connecting the dots and seeing how breakthroughs in AI might impact nanotechnology, or vice versa.'' {\textit{--Yuval Noah Harari, Homo Deus, 2016}}
 \end{quotation}
 
% (a) Are mechanism classes (Direct vs. Indirect) important for our last experiments of search quality? if yes, mention it... In all experiments, we have ignored the class label and just focused on identifying relations.  -- \tom{yes, added}

The effort to mitigate the COVID-19 pandemic is an  interdisciplinary endeavor the world has rarely seen~\cite{nytimes}. As one recent example, expertise in virology, physics, epidemiology and engineering enabled a group of 200 scientists to understand and bring attention to the airborne transmissibility of the SARS-CoV-2 virus \cite{indoor}. The diverse and rapidly expanding body of past and present findings related to COVID-19 \cite{wang2020cord} makes it challenging to keep up, hindering scientists' pace in making new discoveries and connections.

%\dan{good so far, but it would be even stronger if we have evidence (or a citations) to say that MECHANISMS are the key missing link. WHY are we so sure that this is the answer (over, say, broad openIE? Maybe Kuipers comes here?}

\begin{figure}
\centering
\includegraphics[width=\columnwidth]{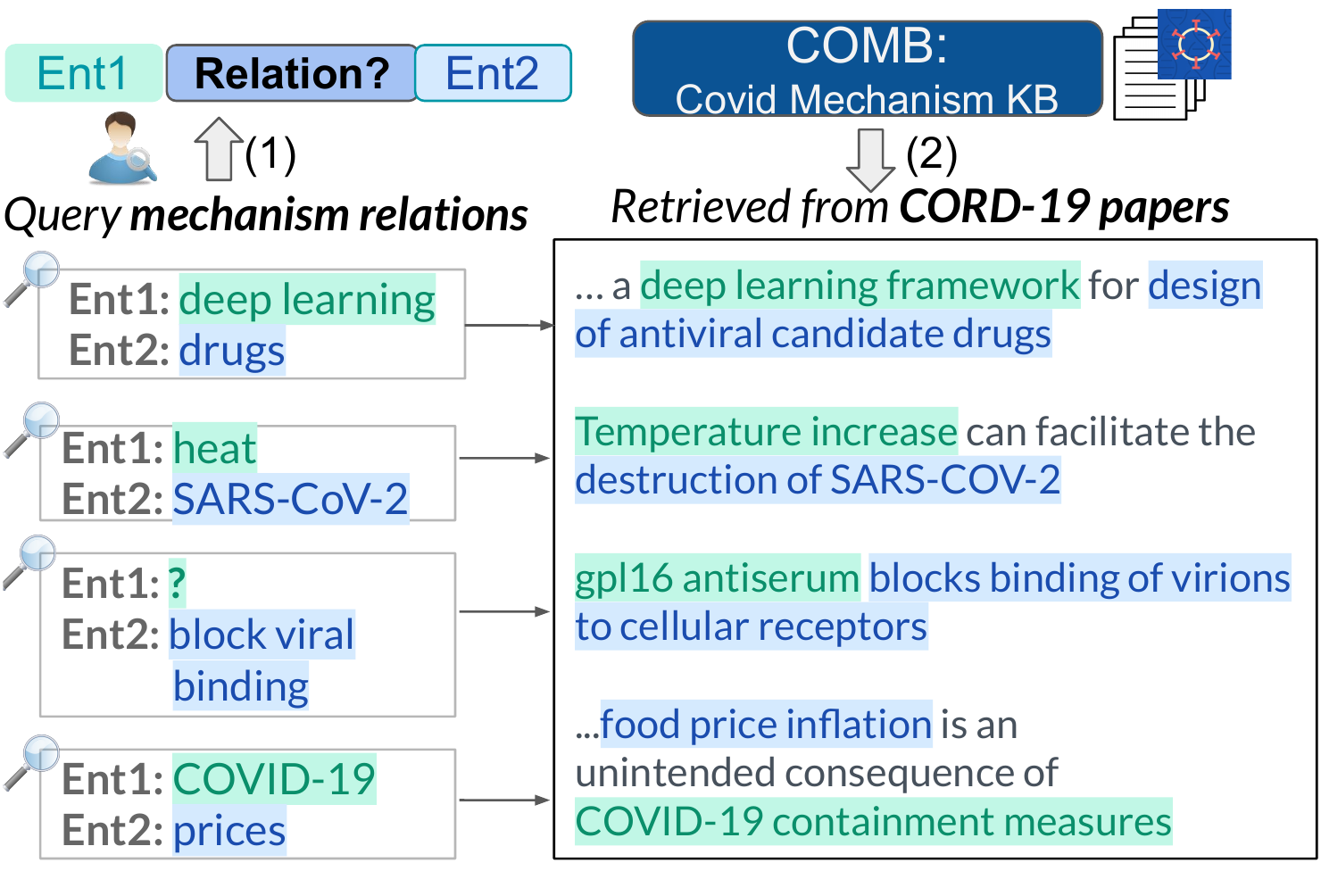}  
\caption{Our COVID-19 Mechanism KB (\kb) is extracted from scientific papers and can be searched for diverse activities, functions and influences (1), retrieving relations from the literature (2).} 
    \label{fig:teaser}
\end{figure}

Research in natural language processing (NLP) has provided important resources to extract {\it fine-grained} relations from scientific papers in specific areas, such as certain subfields of biomedicine  ~\cite{kim2013genia,nye2018corpus} or computer science~\cite{wadden2019entity}. However, these cover only a fraction of all concepts in the literature; in biomedicine alone, there are myriad concepts \cite{salvadores2013bioportal} not covered by NLP resources. For COVID-19 research, the challenge is especially pronounced due to diversity and emerging concepts; even reading just one paper may require background knowledge in multiple biomedical subfields, physics, chemistry, engineering, computer science and the social sciences. %\tnote{ "looks good, so added to text" - dan
For example, consider a paper studying the indoor dynamics of aerosolized SARS-CoV-2 and the effect of ventilation on transmission by using simulation models, or work on economic impacts of COVID-19 on prices and consumption.

To make progress in consolidating such diverse information, we introduce a unified schema of \emph{mechanisms} as a \emph{unified language} covering activities, functions and influences across the sciences. These can be proteins that block viral binding, algorithms to design drugs, the effect heat has on viruses, or COVID-19 has on food prices (Fig.~\ref{fig:teaser}). 

We build on the fact that mechanisms underlie much of the natural language of scientific papers \cite{rohl2012mechanisms}, and construct a unified schema with two coarse-grained mechanism relations:
\begin{itemize}[noitemsep,topsep=.5mm]
    \item \textit{Direct Mechanisms}: mechanistic \textit{activities} (e.g., viral binding) or \textit{functions} engendered by natural or artificial entities (e.g., a protein used for binding or algorithm used for diagnosis).
    \item \textit{Indirect Mechanisms}: \textit{influences and associations} such as economic effects of COVID-19 or complications associated with medical procedures.
\end{itemize}

Our coarse-grained relation schema, over free-form text spans, strikes a balance between the granular information extracted by Closed-IE approaches~\cite{Freitag1998TowardGL,Hoffmann2010Learning5R} and the schema-free breadth of Open IE approaches~\cite{etzioni2008open,stanovsky2018supervised}, which often lead to generic and uninformative relations for scientific applications \cite{Kruiper2020_SORE}.

Furthermore, our schema facilitates %the annotation of a diverse corpus and the 
construction of a high-quality KB that synthesizes interdisciplinary knowledge.
We construct precisely this, releasing \data (\textbf{Mech}anisms \textbf{AN}otated in \textbf{C}OVID-19 papers) -- an annotated dataset of 2,400 mechanisms based on our schema. We  train a state-of-the-art model to extract this information from scientific papers, and use it to build  \kb (\textbf{C}OVID-19 \textbf{O}pen \textbf{M}echanism Knowledge \textbf{B}ase) --  a broad-coverage KB of 1.5M mechanisms in COVID-19 papers. We analyze the characteristics of \kb, showing the distribution of relations across scientific subfields and comparing their quality to other IE approaches.

We demonstrate the utility of \kb in two studies with experts. In the first study, our system achieves high precision and recall in scientific search with structured queries on both diverse viral mechanisms and applications of AI in the literature. In the second study, we evaluate \kb in a usability study with MDs active in treating and researching COVID-19. Our system is rated higher than PubMed search by the clinical experts, in terms of utility and quality.
% We lay the foundations for the framework, which we hope will open new avenues for knowledge representation and discovery across the sciences.

\noindent \textbf{Our main contributions include:}
\begin{itemize}[topsep=1mm,noitemsep]
    \item We introduce a unified schema for {\em mechanisms} that generalizes across many types of activities, functions and influences. We construct and distribute \data, an annotated dataset of papers related to COVID-19, with 2,400 instances of our mechanism relation. %based on this schema.
    \item Using \data, we train an IE model and apply it to 160K abstracts in COVID-19 literature, constructing \kb, a KB of 1.5M mechanism instances. Manual evaluation of relations sampled from our KB shows them to have 88\% accuracy. We also find a model trained on our data reaches roughly 80\% accuracy on a sample of general biomedical papers from across the PubMed corpus, with no additional training, demonstrating the generalization of our approach.
     \item We showcase the utility of \kb in structured search for mechanisms in the literature. In a study with MDs working to combat COVID-19, our system is rated higher than PubMed search in terms of utility and quality. 
      
\end{itemize}

\section{Related work}
\label{sec:mechie}

% In Sections \ref{sec:kbc} and \ref{sec:kbeval}, we employ this schema to construct a KB with an IE approach.which is later used to define in more detail the schema of our task.

%In our work we aim to achieve broad coverage of mechanism relations, extending to a wide range of entities and topics observed in COVID-19 papers. For example, in addition to areas such as medicine, microbiology, genetics, proteomics, zoology and virology, topics we cover in our mechanism annotations include computer science, public policies, flow dynamics, building engineering, macroeconomic impacts and international relations.
% and collecting an annotated dataset of scientific texts from the CORD-19 corpus to train an information extraction model that we can apply to all papers in the corpus.

% \tnote{mechanisms unify}
% \tnote{intro -- why do we need mechanisms???}
% \tnote{emphasize the unified language, underlying language shared between }

% \tnote{small table with detailed examples, and/or start with more basic things make it easier for nlp readers}
% \tnote{explain abstract/natural entities}
% \tnote{abstract of guideline points -- how did we define mechanisms?}

\begin{table*}[t]
\setlength{\belowcaptionskip}{-0pt} 
{\small
\begin{tabular}{p{0.08\linewidth}p{0.205\linewidth}p{0.21\linewidth}p{0.4\linewidth}}
    \toprule
    \textbf{Schema} &
        \textbf{Entity types} &
    \textbf{Relations } & 
    \textbf{Example} \\
    \midrule 
      \parbox{\linewidth}{SciERC}%\cite{luan2018multi}}
        &\parbox{\linewidth}{CS methods/tasks \newline (free-form spans) }    
    &\parbox{\linewidth}{used-for}  
    &\parbox{\linewidth}{Use \gspan{GNNs}for \gspan{relation extraction.}}\\
    \midrule 
     \parbox{\linewidth}{SemRep}% \cite{kilicoglu2011constructing}}
         &\parbox{\linewidth}{Clinical (drugs, diseases, anatomy \dots)}    
    &\parbox{\linewidth}{causes, affects, treats, \newline inhibits, interacts, used \dots}  
    &\parbox{\linewidth}{\dots intratympanic \gspan{dexamethasone injections}for patients with intractable \gspan{Meniere's disease.}}\\
        \midrule
    \parbox{\linewidth}{ChemProt}%\cite{li2016biocreative}}
        &\parbox{\linewidth}{Chemicals, proteins}    
    &\parbox{\linewidth}{direct/indirect regulator, \\ inhibitor, activator \dots}  
    &\parbox{\linewidth}{\gspan{Captopril}inhibited \gspan{MMP-9}expressions in right ventricles.}\\
    \midrule 
    \parbox{\linewidth}{DDI}%\cite{luan2018multi}}
        &\parbox{\linewidth}{Drugs}    
    &\parbox{\linewidth}{interacts}  
    &\parbox{\linewidth}{\gspan{Quinolones}may enhance the effect of \gspan{Warfarin.}}\\
    \midrule 
    \parbox{\linewidth}{GENIA}%\cite{luan2018multi}}
        &\parbox{\linewidth}{Proteins, cellular entities }    
    &\parbox{\linewidth}{binding, modification, \\ regulation \dots}  
    &\parbox{\linewidth}{\gspan{BMP-6}induced phosphorylation of \gspan{Smad1/5/8.}}\\
    \midrule 
        \parbox{\linewidth}{PICO}%\cite{luan2018multi}}
        &\parbox{\linewidth}{Clinical}    
    &\parbox{\linewidth}{Interventions, outcomes}  
    &\parbox{\linewidth}{The \gspan{bestatin} group achieved \gspan{longer remission.}}\\
    \midrule 
  
        \parbox{\linewidth}{\textbf{Ours: \\ \data} }%\cite{luan2018multi}}
        &\parbox{\linewidth}{Medicine, epidemiology, genetics, molecular bio., CS, math, ecology, \newline economics \dots (free-form)}    
    &\parbox{\linewidth}{direct (activities, functions) / indirect (influences, \\ associations)}  
    &\parbox{\linewidth}{$\cdot$ \gspan{RL} can be used to learn \gspan{mitigation policies in epidemiological models.} \newline
     $\cdot$ \gspan{Histophilus-somni} causes \gspan{respiratory, reproductive, cardiac and neuronal diseases in cattle.}} \\
    \bottomrule 
\end{tabular}
}
\caption{Our broad concept of mechanisms covers many relations within existing science-IE schemas. The table shows examples of representative schemas, and the types of entities and relations they capture.}
\label{tab:ieexamples}
\end{table*}

\noindent \textbf{Mechanisms in science} The concept of \emph{mechanisms}, also referred to as \emph{functional relations}, is fundamental across the sciences. For example mechanisms are described in biomedical ontologies \cite{burek2006top,rohl2012mechanisms,keeling2019philosophy}, engineering \cite{hirtz2002functional}, and across science.  Mechanisms can be {natural} (e.g., the mechanism by which amylase in saliva breaks down starch into sugar), {artificial} (electronic devices), {non-physical constructs} (algorithms, economic policies), and very often a blend (a pacemaker regulating the beating of a heart through electricity and AI algorithms). 

Although seemingly intuitive, exact definitions of mechanisms are subject to debate in the philosophy of science \cite{rohl2012mechanisms,keeling2019philosophy}. An Oxford dictionary definition of mechanisms refers to \textit{a natural or established process by which something takes place or is brought about}. More intricate definitions discuss ``complex systems producing a behavior'', ``entities and activities productive of regular changes'', ``a structure performing a function in virtue of its parts and operations'', or the distinction between ``correlative property changes'' and ``activity determining how a correlative change is achieved'' \cite{rohl2012mechanisms}.

Abstract definitions can help with generalization across many important types of mechanisms. The schema we propose (Sec.~\ref{sec:schema}) is inspired by such definitions, operationalizing them and making them more concrete, and also simple enough for models and human annotators to identify.

\vspace{.1cm} \noindent \textbf{Information extraction from scientific texts} There is a large body of literature on extracting information from scientific papers, primarily in the biomedical sphere. This information often corresponds to very \textit{specific} types of mechanisms, as shown in Tab.~\ref{tab:ieexamples}. Examples include ChemProt \cite{li2016biocreative} with mechanisms of chemical-protein regulation, drug interactions in the DDI dataset \cite{segura2013semeval}, genetic and cellular activities/functions in GENIA \cite{kim2013genia}, semantic roles of clinical entities \cite{kilicoglu2011constructing}, PICO interventions and outcomes \cite{wallace2016extracting,nye2018corpus}, and computer science methods/tasks in SciERC \cite{luan2018multi}. Such schemas have been used, for example, to extract genomic KBs \cite{poon2014literome} and automate systematic reviews \cite{nye2020understanding}. %These efforts focus on extracting important information in specific domains.
Our schema draws on these approaches, but with a much broader reach across concepts seen in COVID-19 papers (Tab. \ref{tab:ieexamples}, Fig. \ref{fig:mechtop}).

An important area in information extraction focuses on \emph{open} concepts, with prominent approaches being Open IE \cite{etzioni2008open} and Semantic Role Labeling (SRL;  \citealp{carreras2005srl}), which share similar properties and predictions \cite{stanovsky2018supervised}. While such methods are intended to be domain independent, they perform significantly worse in the scientific domain \cite{groth2018open}. \citet{Kruiper2020_SORE} developed a multi-stage process to post-process Open IE outputs, involving trained models and humans to find a balance between generic and fine-grained clusters of relation arguments and omitting noisy clusters. 
% train a model to detect relations indicating ``trade-offs'' in the biology domain. The model is then used to filter open IE relations to be more relevant/informative. This is done in a multi-stage process involving human tuning to find a balance between generic and fine-grained clusters of relation arguments and omitting noisy clusters.  
%
In contrast, our unified schema enables annotating a dataset of mechanism relations between free-form spans and training IE models to automatically generalize across diverse relation types. 

Our schema is also related broadly to the task of training reading comprehension models on procedural texts describing scientific processes (such as short paragraphs written by crowd workers to explain photosynthesis in simple language; \citealp{mishra2018tracking}). Our representation of scientific texts in terms of a graph of causal relations can potentially help infer processes across science.

% \dan{We should connect our work to some of the ARisto work on processes, eg EMNLP18 "Reasoning about Actions and State Changes by Injecting Commonsense Knowledge" or Aida's
%  paper on Procedural Reading Comprehension with Attribute-Aware Context Flow, or Everything Happens for a Reason: Discovering the Purpose of Actions in Procedural Text?
% }

%Our primary goal in this paper is to construct a KB of diverse mechanisms appearing in the CORD-19 corpus. We formulate a novel relation schema that is able to naturally generalize across diverse topics by targeting a broad and fundamental class of relations reflecting activities, functions and influences, between free-form spans.

%Indeed, even within biomedical fields the coverage of the full spectrum of concepts is limited\footnote{The BioPortal repository \cite{salvadores2013bioportal} lists over \textit{12M} classes from over 900 ontologies}, and emerging domains such as COVID-19 further introduce novel concepts.

\vspace{.1cm} \noindent \textbf{COVID-19 IE} Recent work \cite{verspoor2020proceedings} has focused on extracting information from the CORD-19 corpus \cite{wang2020cord}. PICO concepts are extracted and visualized in an exploratory interface in the COVID-SEE system \cite{verspoor2020covid}. 
In \citet{wang2020covid}, genes, diseases, chemicals and organisms are extracted and linked to existing biomedical KBs with information such as gene-disease relations. Additional relations based on the GENIA schema are extracted from the text. To address the novel COVID-19 domain, the schema is enriched with new entity types such as viral proteins and immune responses. 

In this paper, we focus on a more general schema that captures diverse concepts appearing in literature related to COVID-19, an emerging domain with novel concepts coming from many fields and subfields. The mechanism KG we construct includes---as a subset ---diverse biomolecular and clinical information (such as chemical-disease relations) as part of a general mechanism schema.

\label{sec:data}

\section{Mechanism Relation Schema}
\label{sec:schema}

%As discussed in Section \ref{sec:mechie}, the concept of mechanisms is pervasive across the sciences. 
We present a schema that builds upon and consolidates many of the types of mechanisms discussed in Sec.~\ref{sec:mechie}. Our defined schema has three key properties:  (1)~it uses a generalized concept of mechanism relations, capturing specific types of mechanisms in existing schema and extending them broadly; (2)~it includes flexible, generic entities not limited to predefined types, and (3)~it is simple enough for human annotators and models to identify in the natural language of scientific texts. This schema enables forming our KB by identifying a set of mechanism relations in a corpus of scientific documents (Sec.~\ref{sec:KB}).

\begin{figure}[t]
\includegraphics[width=1.0\columnwidth]{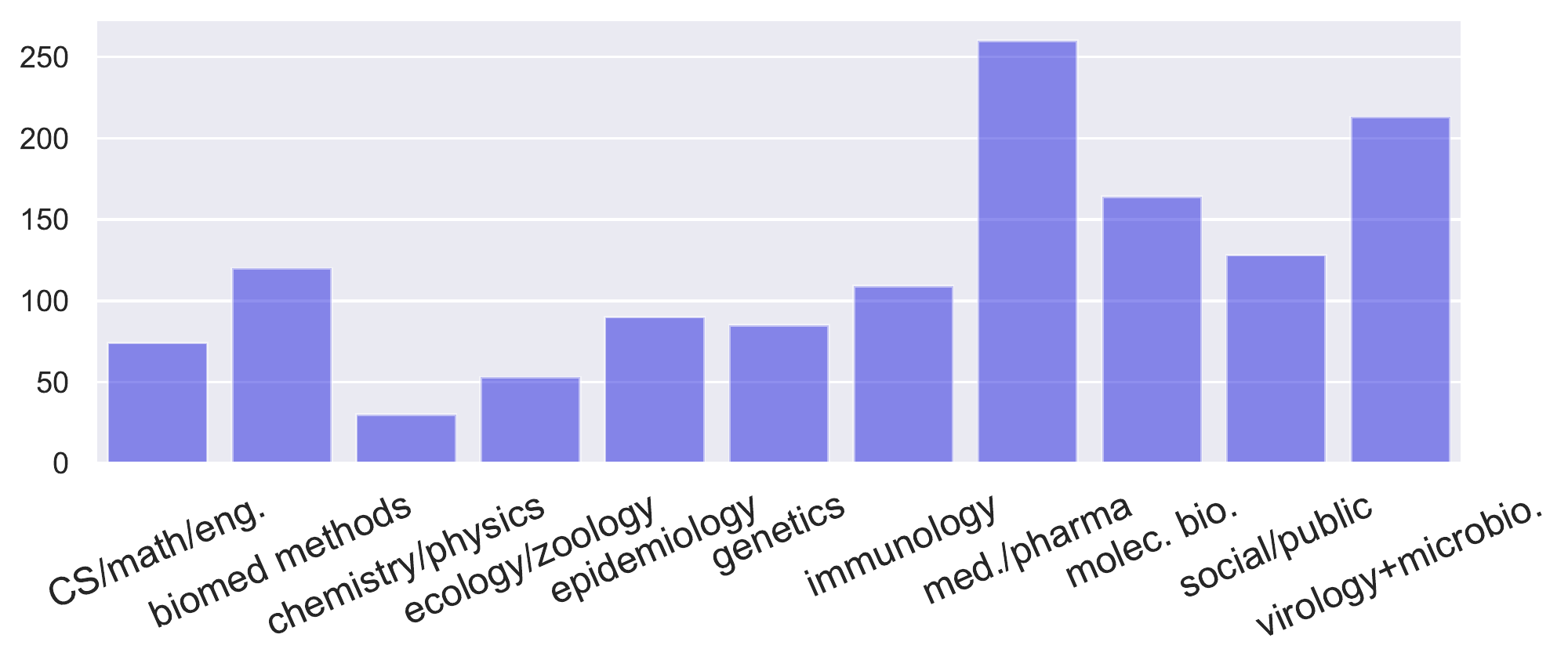}  
\caption{\data covers a diverse set of scientific fields. Histogram of domains in \data (sample of 350 relations). Manually labeled relation entities, based on a list of scientific disciplines from Wikipedia.}
    \label{fig:mechtop}
\end{figure}

We formally define each mechanism as a relation $(E_1,E_2,\text{{\tt class}})$ between entities $E_1$ and $E_2$, where each entity $E$ is a text span and the {\tt class} indicates the type of the mechanism relation. Entities all share a single common type and can be either natural (e.g., protein functions, viral mechanistic activities) or artificial (e.g., algorithms, devices), to capture the generality of the concepts in science (see Fig.~\ref{fig:mechtop}). We allow each entity to take part in multiple relations (tuples) within a given text, leading to a ``mechanism graph''. Mechanisms are categorized into two coarse-grained classes:\footnote{We also provide a dataset and extraction model for ternary relations in the form of \emph{(subject, object, predicate)}. We focus on the coarse-grained mechanism schema due its broader flexibility and coverage. See App.~\ref{appendix:granular} for details.}

\vspace{.1cm}\noindent{\bf Direct mechanisms} include {\it activities} of a mechanistic nature -- actions explicitly performed by an entity, such as descriptions of a virus binding to a cell, and explicit references to a function (e.g., a use of a drug for treatment, or the use of AI for drug design as in Fig.~\ref{fig:teaser}).

\vspace{.1cm}\noindent{\bf Indirect mechanisms} include influences or associations without explicit mechanistic information or mention of a function (such as describing observed effects, without 
the process involved). These relations correspond more to ``input-output correlations'' \cite{rohl2012mechanisms}, such as indicating that COVID-19 may lead to economic impacts but not \emph{how} (Fig.~\ref{fig:teaser}), as opposed to direct mechanisms describing ``inner workings'' -- revealing more of the intermediate states that lead from initial conditions (COVID-19) to final states (price inflation) or explicitly describing a function. As an example for the utility of this distinction between direct and indirect relations, consider an MD looking to generate a structured list of all \textit{uses} of a treatment (direct mechanism), but not include side effects or complications (indirect).

\section{KB Construction}
\label{sec:kbc}

\begin{figure*}[t]
\includegraphics[width=\linewidth]{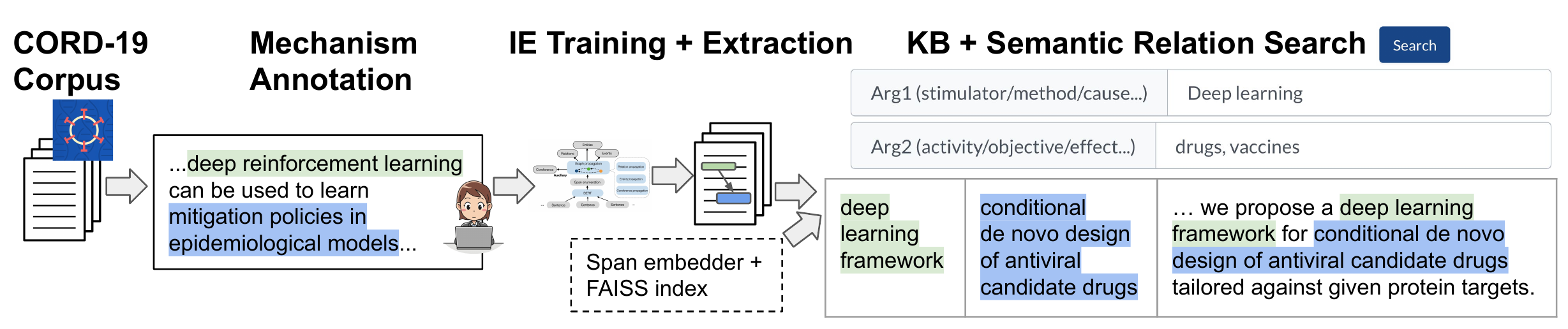}
    \caption{\textbf{Overview of our approach}. We collect annotations of mechanisms (textual relations) from the CORD-19 corpus, which are used to train an IE model. We apply the model to over 160K documents in the corpus, extracting over 1.5M relations that are fed into our KB. Entity mention spans are embedded with a language model tuned for semantic similarity, and indexed with FAISS for fast similarity search as part of our search interface.}
    \label{fig:flow}
\end{figure*}

We describe our approach (depicted in Fig.~\ref{fig:flow}) for extracting  a knowledge base of mechanisms using our unified schema.  We first curate \data, an annotated dataset of general mechanisms from a small collection of scientific papers (Sec.~\ref{sec:data}). We then train a model on our annotated data to extract mechanism relations from the entire CORD-19 corpus of scientific papers; we use it to build \kb, a knowledge base of mechanisms across the entire CORD-19 corpus of (Sec.~\ref{sec:kb_mechanisms}), which supports semantic search for relations (Sec.~\ref{sec:KB}). 

%  We curate a dataset for training because other datasets are domain-specific and unsupervside methods won't work; unsupervised models are not so good (4.1.1)  Annotating a dataset of mechanisms, (4.1.2) Training a model to extract mechanisms from scientific articles, and (4.2) Corpus-level KB and Search: Applying the trained model to extract mechanisms from a large collection of documents and algorithms for searching in this KB.

\subsection{Collecting Mechanism Annotations}
\label{sec:data}

%To be able to extract \emph{general} mechanism relations from across the literature connected to COVID-19 and construct a KB, 
We construct a dataset of mechanism relations in texts randomly sampled from the CORD-19 corpus \cite{wang2020cord} that includes scientific papers connected to COVID-19. To circumvent annotation challenges in scientific datasets~\cite{luan2018multi} and ensure high-quality annotations, we follow a three-stage process of (1) annotating entities and relations using biomedical experts, (2) unifying span boundaries with an NLP expert, and (3) verifying annotations with a bio-NLP expert. Our annotation process is a relatively low-resource and generalizable approach for a rapid response to the COVID-19 emergency.

In the first stage, five annotators with biomedical and engineering background annotate all mechanism relations as defined in Sec.~\ref{sec:schema} (full annotation guidelines are available in our code repository). Relations are annotated as either direct/indirect. Entities are annotated as the longest span of text that is involved in a relation with another entity, while not including redundant or irrelevant tokens. As in related tasks \cite{luan2018multi}, annotators are guided to resolve doubt on span boundaries by selecting the longest relevant span. 

Annotators had a one-hour training session. In the first part of the training session, annotation guidelines were reviewed. The guidelines included simple explanations of direct/indirect mechanisms along with introductory examples (e.g., ``\emph{the virus makes use of spike protein to bind to a cell}'', ``\emph{A virus leads to respiratory infection}''). In the second part, annotators saw examples from papers in the annotation interface (see Fig.~\ref{fig:annotations},  App.~\ref{appendix:data_anno}), and performed a few live training annotations.  % only when involved in a relation with another
%using Prodigy, a platform with a GUI for rapid annotations \cite{Prodigy:2018}. Entities are annotated only when involved in a relation with another. Following \cite{luan2018multi}, annotators were instructed to perform a greedy annotation preferring longer spans whenever ambiguity occurs as to span boundaries. 
%
%We initially observed large variation between annotators and low agreement as measured with strict, exact matching criteria between relations. A deeper look revealed much of the disagreement was due to variations in annotation style rather than meaning. 

We initially observed significant variation between annotators in identifying span boundaries for entity annotations, stemming from inherent subjectivity in such annotation tasks \cite{stanovsky2018supervised,luan2018multi} and from lack of NLP experience by some annotators.
 %This likely stems from the challenging nature of our task with abstract, soft definitions of relations between free-form spans (see examples in Table~\ref{table:annotation_errors}, Appendix~\ref{appendix:IE_eval}). 
In the second stage, an NLP expert annotator conducted a round of style unification by viewing annotations and adjusting span boundaries to be more cohesive while preserving the original meaning, focusing on boundaries that capture essential but not redundant or generic information (e.g., adjusting the span \textit{substantial \textbf{virus replication} by unknown mechanisms} to include only \textit{virus replication}).  
Finally, in the third stage, a bio-NLP expert with experience in annotating scientific papers verified the annotations and corrected them as needed. The expert accepted $81\%$ of the annotations from the second stage without modification, confirming the high quality of the stage-2 data. Relation label mismatches accounted for 5\% of the remaining 19\%. Other sources of disagreement were span mismatches and new relations added by the bio-NLP expert adjudicator.

The resulting dataset (\data: \textbf{Mech}anisms \textbf{AN}otated in \textbf{C}OVID-19 papers) contains 2,370 relation instances (1645 direct, 725 indirect) appearing in 1,000 sentences from 250 abstracts.\footnote{The dataset is similar in size to related scientific IE datasets \cite{luan2018multi} which share related challenges in collecting expert annotations of complex or ambiguous concepts over difficult texts.} Average span length is 4 tokens, while the average distance between relation arguments is 11.40 tokens. 

\subsection{Extracting a KB of Mechanisms} 
\label{sec:kb_mechanisms}
Using \data, we train an IE model to extract mechanism relations from sentences in scientific documents. We train  DyGIE++~\cite{wadden2019entity}, a state-of-the-art end-to-end IE model which extracts entities and relations jointly (without assuming to have entity spans given), classifying each relation as one of $\{\text{{\tt DIRECT}},\text{{\tt INDIRECT}}\}$.\footnote{We use DyGIE++  with SciBERT \cite{beltagy-etal-2019-scibert}  embeddings fine-tuned on our task and perform  hyperparameter grid search (for dropout and learning rate only) and select the best-performing model on the development set ($7e-4$ and $0.43$, respectively). Full details are in App.~\ref{appendix:hyperparam}.} 

To form our corpus-level KB, we apply the trained model to each document in our corpus (all 160K abstracts in the CORD-19 corpus) to extract mechanism relations  and then integrate the extracted relations. We find that our trained model achieves high precision scores for high confidence predictions  (precision $\geq 80\%$ within top-$20$ predicted relations; see $P@K$ figure, App.~\ref{appendix:IE_eval}). Therefore, our corpus-level KB is constructed by filtering predictions with low confidence. %The extracted high-confidence mechanisms  include relations  %$r_{i} \in \mathcal{M}$ 
%as  tuples in the form of $(E_1,E_2,\text{{\tt DIRECT/INDIRECT}})$, where entities $E_i$ are free-form spans of text. 
% Full details are provided in Appendix~\ref{appendix:IE_eval}.

To integrate relations and entities across the corpus, we use standard surface-level string normalization (such as removing punctuation, lemmatizing, and lowercasing) and unify and normalize entity mentions using coreference clusters of entities within a document.\footnote{We use a pre-trained DyGIE++ model trained on SciERC to obtain coreference clusters.} Each coreference cluster is assigned a representative entity as the mention with the longest span of text, and all other entities in that cluster are replaced with the representative entity. This  is particularly useful for normalizing pronouns such as \textit{it} with the original mention they referred to (e.g., a specific virus or method \textit{it} refers to). 

Our final KB (\kb) consists of 1.5M relations in the form of  $(E_1,E_2,\text{{\tt DIRECT/INDIRECT}})$ filtered by high confidence score ($>= 90\%$), where entities $E_i$ are standardized free-form spans of text.

\subsection{Semantic Relation Search} \label{sec:KB} The constructed KB enables applications for  retrieving relations across concepts from many disciplines. For example, searching for  all documents that include  mechanisms to incorporate {\it AI} in studies of {\it heart disease} $(E_1 = \text{AI},E_2 = \text{heart disease},\text{{\tt DIRECT}})$ requires going beyond simply finding documents that mention  \textit{AI} and \textit{heart disease}. Here, we describe our approach for searching over the KB by encoding entities and relations,  capturing related concepts (such as \emph{cardiac disease} and \emph{heart conditions}), as well as simpler surface matches (\emph{artificial intelligence \textbf{methods}}, \emph{artificial intelligence \textbf{models}}).

Specifically, for a given query $\mathbf{q} \coloneqq (E^q_{1},E^q_{2}, \text{{\tt class}})$, our goal is to find  mechanisms $r_i$ in \kb whose entities are free-form texts similar to  $E^q_{1},E^q_{2}$ in the query. The {\tt class} is used to filter for the type of relation---for example, when explicitly requiring {\tt DIRECT} mechanisms.

%. we define queries to be $\mathbf{q} \coloneqq (E^q_{1},E^q_{2}, \text{{\tt class}})$, where $E^q_{1}$ and $E^q_{2}$ are free-form texts corresponding to the first and second entities in a mechanism relation, and $\text{{\tt class}} \in \{\text{{\tt DIRECT}},\text{{\tt INDIRECT}}\}$, respectively.  

\vspace{0.2cm}\noindent\textbf{Entity encoding} We obtain an encoding function $f:E\mapsto\mathbb{R}^d$ to encode all unique spans (entities) in the KB to a $d$ dimensional vector space. The encoding function is derived by fine-tuning a language model (LM) originally trained on PubMed papers~\cite{Gururangan2020DontSP} on semantic similarity tasks. For fine-tuning, we use sentence pairs in STS \cite{cer2017semeval} and SNLI \cite{bowman2015large}  following~\citet{reimers-2019-sentence-bert}, and add  biomedical sentence pairs from the BIOSSES dataset \cite{souganciouglu2017biosses}.

\vspace{0.2cm}\noindent\textbf{Relation similarity} Given a query $\mathbf{q}$, we rank the set of all \kb relations with the same  {\tt class} as the query. For each candidate relation $r=(E_1,E_2,\tt class)$ in \kb, we compute its similarity to the query relation $\mathbf{q}$  as the minimum similarity between encodings of their corresponding entities: $\min\limits_{j \in \{1,2\}} f(E_{j}) \cdot f(E^q_{j})$. With this definition, a relation $(E_1,E_2)$ with $E_1$ very similar to the first entity of the query $E^q_{1}$ but $E_2$ distant from $E^q_{2}$ will be ranked low. For example, with the query $(E^q_{1}=\text{deep learning},E^q_{2}=\text{drugs})$, the relation $(E_1=\text{microscope},E_2=\text{drugs})$ will be ranked low due to the pair (deep learning, microscope).
 For efficient search, we create an index of embeddings corresponding to the 900K unique surface forms in \kb and employ a system designed for fast similarity-based search \cite{JDH17}.

\begin{figure*}[h]
\begin{subfigure}{.3\textwidth}
\includegraphics[width=.9\linewidth]{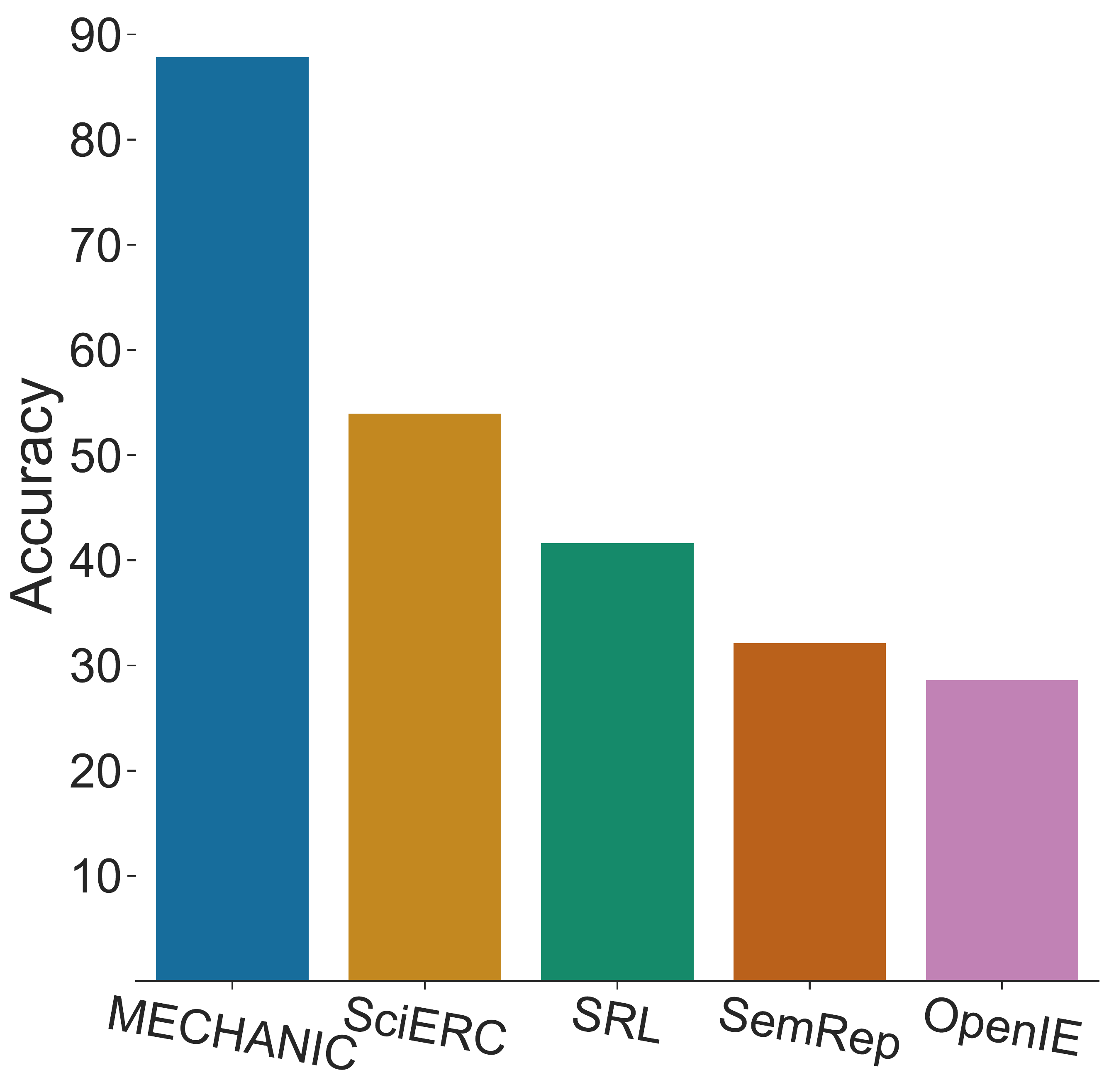}
\end{subfigure}
\begin{subfigure}{.33\textwidth}
\includegraphics[width=.9\linewidth]{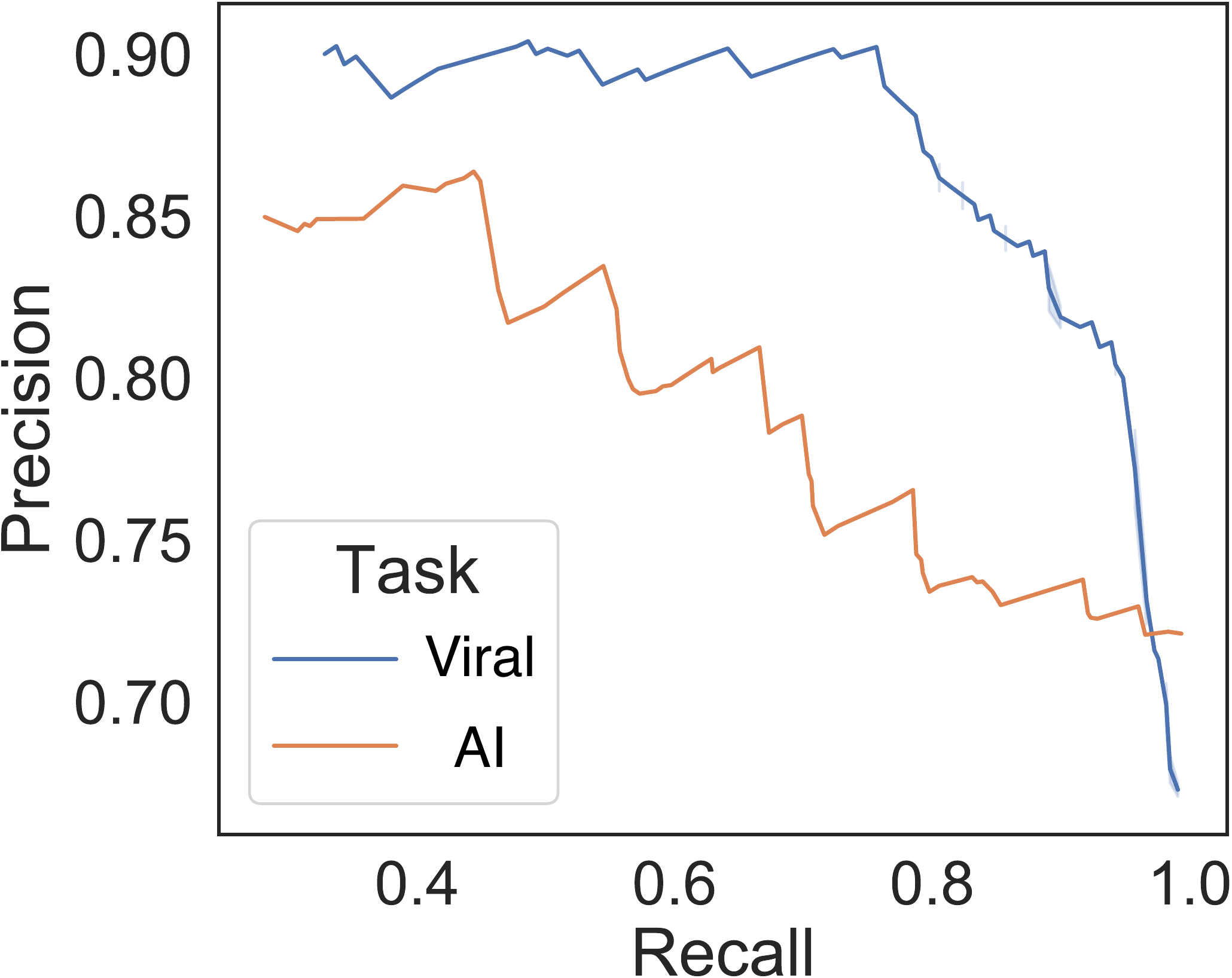} \vspace{-.2cm}
\end{subfigure}
\begin{subfigure}{.35\textwidth}
\begin{tabular}{|c|c|c|} 
 \hline
 Metric & PubMed & \kb \\ 
 \hline
 Search  & 71\%  & \textbf{90\%} \\
 Utility  & 69.5\%  & \textbf{92\%} \\
 Interface  & 78\%  & \textbf{90\%} \\
 Overall  & 74\%  & \textbf{91\%} \\
 \hline
 \end{tabular} 
% \end{small}
\vspace{-.3cm} \end{subfigure}
    \caption{Evaluating \kb in studies with experts. \kb is found to have high quality and utility, outperforming other approaches. \textbf{Left:} \kb outperforms  external resources with either \textit{specific} types of mechanisms or \textit{open} relations, in human evaluation for correctness and usefulness of predictions, on a sample of 300 predicted relations. \textbf{Center:} Retrieved relations are ranked by query similarity (Sec.~\ref{sec:KB}) and compared to human relevance labels to compute precision/recall. Higher-ranked results are overall judged as relevant by humans. \textbf{Right:} Results of \kb search study with five practicing MDs, using both our system and PubMed to search the literature. Experts were given a post-study questionnaire with questions grouped by subject (search, utility, interface). Our mechanism search system performed substantially better than PubMed.
}
    \label{fig:kb_eval}
\end{figure*}

\section{Evaluating \kb} 
\label{sec:kbeval}

In this section, we evaluate the constructed KB of mechanisms in terms of correctness and informativeness (Sec.~\ref{sec:kb_eval_corr}), and its utility in searching for mechanisms (Sec.~\ref{sec:kb_eval_util}). Our main goal is to ensure the mechanism relations have high quality to support our large-scale KB and search applications. We further show that our schema is useful as compared to other schema.

\begin{table*}[t]
\begin{subtable}[t]{\textwidth}
\centering

{\small
\begin{tabular}{|p{0.23\linewidth}|p{0.45\linewidth}|}
\hline
   \textbf{Relation query} &
    \textbf{Example results from KB search interface}\\
 \hline
     \parbox{\linewidth}{\begin{minipage}{\linewidth}{\vspace{.2\baselineskip}\includegraphics[height=.9cm]{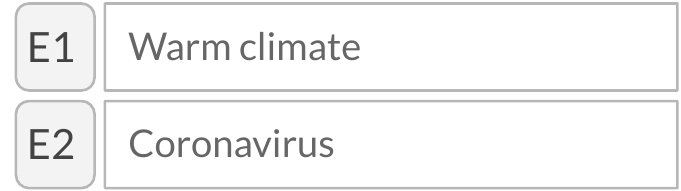}\vspace{.2\baselineskip}}  \end{minipage}}
    &\parbox{\linewidth}{ \begin{minipage}{\linewidth}{\includegraphics[width=\linewidth]{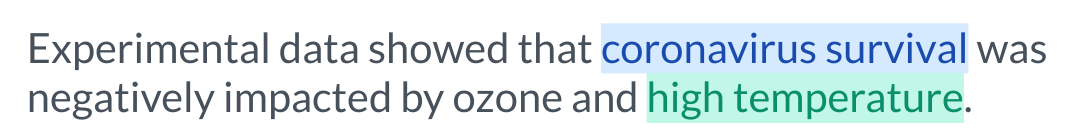}}  \end{minipage}}\\
    \hline
     \parbox{\linewidth}{\begin{minipage}{\linewidth}{\vspace{.2\baselineskip}\includegraphics[height=.9cm]{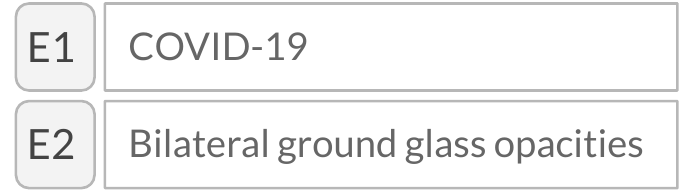}\vspace{.2\baselineskip}}  \end{minipage}}
    &\parbox{\linewidth}{ \begin{minipage}{\linewidth}{\includegraphics[width=\linewidth]{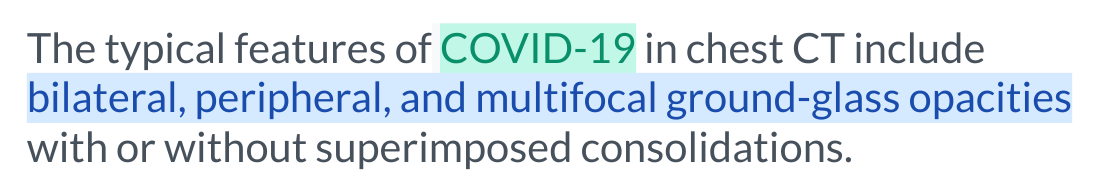}}  \end{minipage}}\\
    \hline
    \parbox{\linewidth}{\begin{minipage}{\linewidth}{\vspace{.2\baselineskip}\includegraphics[height=1cm]{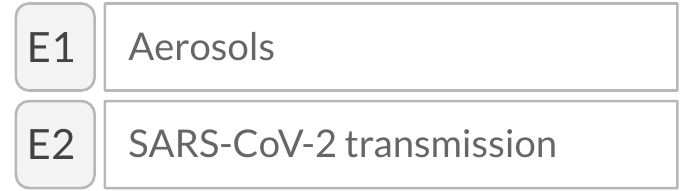}\vspace{.2\baselineskip}}  \end{minipage}}
    &\parbox{\linewidth}{ \begin{minipage}{\linewidth}{\includegraphics[width=\linewidth]{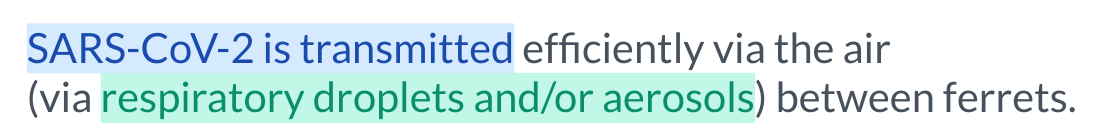}}\end{minipage}}\\
    \hline
\end{tabular}
}
\caption{\textbf{Viral mechanism search}. Queries for $(E1,E2)$ relations, and example retrieved results.}
\label{tab:searchres_ex}
\end{subtable}
\begin{subtable}[t]{\textwidth}
\centering
\vspace{0.25cm}

{\small
\begin{tabular}{|p{0.23\linewidth}|p{0.45\linewidth}|}
    \hline
     \parbox{\linewidth}{\begin{minipage}{\linewidth}{\vspace{.4\baselineskip}\includegraphics[height=1cm]{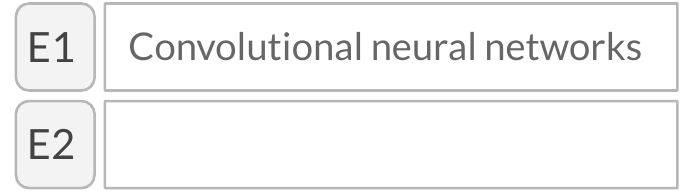}\vspace{.4\baselineskip}}  \end{minipage}}
    &\parbox{\linewidth}{ \begin{minipage}{\linewidth}{\includegraphics[width=\linewidth]{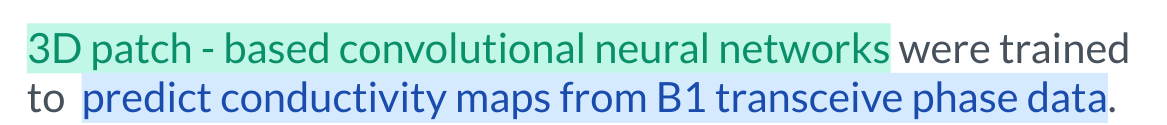}}  \end{minipage}}\\
    \hline
     \parbox{\linewidth}{\begin{minipage}{\linewidth}{\includegraphics[height=1cm]{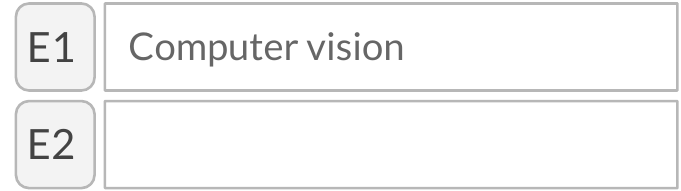}}  \end{minipage}}
    &\parbox{\linewidth}{ \begin{minipage}{\linewidth}{\includegraphics[width=\linewidth]{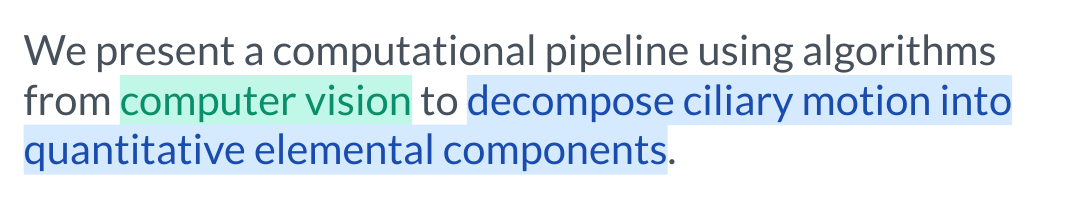}}  \end{minipage}}\\
    \hline
    \parbox{\linewidth}{\begin{minipage}{\linewidth}{\includegraphics[height=1cm]{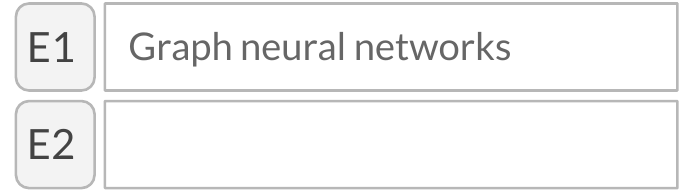}}  \end{minipage}}
    &\parbox{\linewidth}{ \begin{minipage}{\linewidth}{\includegraphics[width=\linewidth]{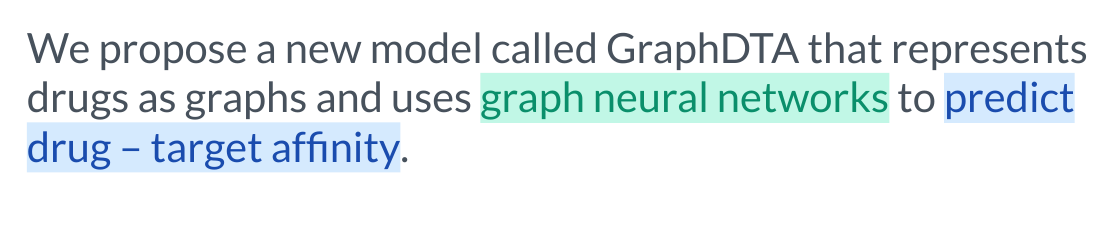}\vspace{-0.5\baselineskip}}  \end{minipage}}\\
    \hline
\end{tabular}
}
\caption{\textbf{AI search}. Queries consists of only $E_1$, to find all applications of AI approaches/areas.}
\label{tab:searchres_ex2}
\end{subtable}
\caption{Example search queries and results for the viral mechanism and AI applications tasks.}
\end{table*}

\subsection{KB Correctness and Informativeness}
\label{sec:kb_eval_corr}
We employ two annotators with biomedical and CS backgrounds to judge the quality of the predicted relations in \kb. In particular, following \citet{groth2018open}, 
annotators are given a predicted relation together with the sentence from which it was extracted.  We collapse all entities/relations into one generic type for this analysis. Annotators are asked to label the predicted relation as correct if (1) it accurately reflects a mechanistic relation mentioned in the sentence (\textit{correctness}), and (2) the extracted entities and relation label are sufficient to convey the meaning of the relation, without referring to the source sentence (\textit{informativeness}). We collect human judgements for 300 predicted relations for our approach and baselines, sampled from 150 randomly selected sentences. Agreement is $71\%$ by Cohen's Kappa and $73\%$ by Matthew's Correlation Coefficient.

% NOTE(dwadden) I replaced this with the version above.
% given predicted relations and corresponding sentences annotators are asked to tag relations as positive if they reflect a mechanism described in the sentence (\textit{correctness}) AND if they can be sufficiently understood by viewing it without the original sentence, consisting of coherent spans that capture essential information but not redundant or irrelevant information (\textit{informativeness}). 

\vspace{.1cm} \noindent {\bf Comparing KB quality to other schemas} To showcase the benefit of our approach, we compare the relations extracted using a DyGIE model trained on \data, versus a DyGIE model trained on other resources that are most related to our mechanisms: SemRep \cite{kilicoglu2011constructing} captures a wide range of biomedical relations (such as drug-drug interactions), and SciERC \cite{luan2018multi} contains relations relevant to computer science (such as ``method-task'' and ``used-for'' relations).\footnote{We use an expert annotator to align external resources to our {\tt direct} or {\tt indirect mechanism} annotations, e.g., USED-FOR is mapped to \textit{direct} mechanism).} In addition, we compare with a Semantic Role Labeling (SRL) method \cite{shi2019simple} that captures broad relations between free-form spans that focus on agents and actions, and a neural OpenIE model \cite{stanovsky2018supervised}.

 Fig.~\ref{fig:kb_eval} (left) shows that $88\%$ of relations from \kb are marked as correct by human raters, demonstrating that our approach extracts mechanism relations with better quality than external resources.\footnote{We also experiment with automated evaluation. We split \data into train/dev/test sets (170/30/50 abstracts), and obtain $F1=50.2$ for entity detection, $F1=45.6$ for relation detection and $F1=42.8$ for classification, on par with performance in other similar scientific IE tasks \cite{luan2018multi}. See more details in App.~\ref{appendix:cofie_model}.} These results suggest that our predicted relations are of overall high quality and can be used to build our corpus-level KB and explore its utility.
% This large gap likely stems from the inability of the automated evaluation metric to reflect the criteria human raters were asked to follow (i.e., rating as positive relations with spans that capture essential information only, and sufficient information to be understood without surrounding context). 

\vspace{.1cm} \noindent {\bf Examining Generalization} COVID-19 papers are highly diverse both topically and chronologically. We conduct a small-scale preliminary experiment examining whether a model trained on \data can generalize to capture mechanism relations in the general biomedical papers, from a much larger corpus of open access papers on PubMed Central (PMC).\footnote{\href{https://www.ncbi.nlm.nih.gov/pmc/}{https://www.ncbi.nlm.nih.gov/pmc/}} We randomly sample a set of 200 predicted relations from papers across the entire PMC corpus, and label them using the same criteria used above. As expected, we find that performance drops, but encouragingly is still considerably high: after filtering predictions with confidence lower than 90\% in the same way we construct \kb, 76\% of relations are considered correct. When filtering for confidence with a threshold of 95\% (which captures 70\% of the samples), the rate of correct predictions is 78\%. In future work it would be interesting to fine-tune our model on a small set of labeled examples from the general PMC corpus to potentially improve these results.

\subsection{\kb Utility} 
\label{sec:kb_eval_util}
We design several search tasks and user studies to evaluate the utility of the constructed KB (Sec.~\ref{sec:searchQuality}) and compare it with the PubMed medical KB and search engine (Sec.~\ref{sec:pubmed}), as judged by medical doctors working on the front lines of COVID-19 treatment and research. All tasks are designed to evaluate our framework's utility in helping researchers and clinicians looking to quickly search for mechanisms or cause-effect relations in the literature and retrieve a list of structured results.

\subsubsection{Search Quality} \label{sec:searchQuality} We form search queries based on a wide range of topics pertaining to (1) SARS-CoV-2 mechanisms (such as modes of transmission, drug effects, climatic influences, molecular-level properties) and (2) applications of AI in this area. Tab.~\ref{tab:searchres_ex} and ~\ref{tab:searchres_ex2} show queries and example relations returned from \kb, along with the context sentences from which they were extracted. 

\vspace{.1cm} \noindent {\bf Viral mechanism search} Queries are formed based on statements in recent scientific claim-verification work (\citealp{wadden2020fact}; see full list in App.~\ref{appendix:kb_util}). For example, for the statement  \emph{the coronavirus cannot thrive in warmer climates}, we form the query as 
$(E_1=\text{Warm climate},E_2=\text{coronavirus})$ (see Tab.~\ref{tab:searchres_ex} row 1). %  and retrieve relations where $E^*_1$ is relevant/similar to ``Warm climate'' and $E^*_2$ relevant to ``COVID-19'' (see Table \ref{tab:searchres_ex}).
For statements reflecting an indirect association/influence, we filter for {\tt INDIRECT} relations (Tab.~\ref{tab:searchres_ex} row 2).  %(e.g., \emph{Bilateral ground glass opacities are often seen on chest imaging in COVID-19 patients}), we filter for {\tt INDIRECT} relations. 
For statements that reflect an undirected mechanism relation (e.g., \emph{Lymphopenia is associated with severe COVID-19 disease}), we query for both directions. %, i.e., both $(E_1=\text{Lymphopenia},E_2=\text{COVID-19})$ and $(E_1=\text{COVID-19},E_2=\text{Lymphopenia})$.

% NOTE(dwadden) old version here.
% The search query only includes the first entity with popular, leading subfields and methods within AI (e.g., \textit{deep reinforcement learning} or \textit{text analysis}. Since all queries relate to \textit{uses} of AI, we filter for {\tt DIRECT} relations. 
\vspace{.1cm} \noindent{\bf AI applications search} % We form queries such that  
This task is designed to explore the uses of AI in COVID-19 papers (Tab.~\ref{tab:searchres_ex2}). We use queries where the first entity $E_1$ is a leading subfield or method within AI (e.g., \textit{deep reinforcement learning} or \textit{text analysis}), and the second entity $E_2$ is left unspecified. Since all queries relate to \textit{uses} of AI, we filter for {\tt DIRECT} relations. These open-ended queries simulate an exploratory search scenario, and can potentially surface inspirations for new applications of AI against COVID-19 or help users discover where AI is being harnessed.

\vspace{.1cm}\noindent {\bf Evaluation} Expert annotators are instructed to judge if a relation is related to the query or not and if the sentence actually expresses the mechanism. These annotations are used as ground-truth labels to compute precision/recall scores of the relations extracted by our algorithm. Since it is not feasible to label every relation, annotators are shown a list of 20 relations for each query including high and low rank relations returned by our search algorithm.\footnote{Specifically, for each query we retrieve the top-1000 similar relations from \kb, ranked as described in Sec.~\ref{sec:kbc}, and select the top and bottom 10 relations (20 per query, 200(=20x10) per task, 400(=200x2) in total), shuffle their order, and present to annotators together with the original sentence from which each relation was extracted.} In total, we use 5 annotators to obtain 1,700 relevance labels across both tasks. Inter-annotator agreement is high by several metrics, ranging from $0.7$--$0.8$ depending on the metric and task; see App.~\ref{appendix:kb_util}. Annotators have graduate/PhD-level background in medicine or biology (for the first task) and CS or biology (for the second task). 

% We ask evaluators to judge each retrieved relation $(E1,E2)$ by whether (1) it is relevant to the query, and (2) the sentence in which $(E_1,E2)$ is mentioned expresses a mechanism relation between the two terms, rather than incidentally mentioning them together. 

\vspace{.1cm} \noindent {\bf Results} Fig.~\ref{fig:kb_eval} (center) shows our results for both tasks. For biomedical search queries, we observe $90\%$ precision that remains stable for recall values as high as $70\%$. For {\it AI applications} we observe a precision of $85\%$ at a recall of $40\%$ that drops more quickly. This lower precision is likely due to the fact that $E_2$ is unspecified, leading to a wider range of results with more variable quality.

Overall, these results showcase the effectiveness of our approach in searching for mechanisms between diverse concepts in COVID-19 papers. 

\subsubsection{Comparing \kb\ with PubMed}
\label{sec:pubmed}
This experiment compares the utility of \kb in structured search for causal relationships of clinical relevance to COVID-19 with PubMed\footnote{\href{https://pubmed.ncbi.nlm.nih.gov/}{https://pubmed.ncbi.nlm.nih.gov/}}---a prominent search engine for biomedical literature that clinicians and researchers frequently peruse as their go-to tool. PubMed allows users to control structure (e.g., with MeSH terms or pharmacological actions), is supported by a KB of biomedical entities used for automatic query expansion, and has many other functions.

\vspace{.1cm} \noindent {\bf Expert evaluation} We recruit five expert MDs---with a wide range of specialities including gastroenterology, cardiology, pulmonary and critical care---who are active in treating COVID-19 patients and in research. Each expert completed search randomly ordered tasks using both PubMed and our \kb UI, showing the full set of ranked relations, as well as the sentence snippet mentioning the relation, the paper title, and hyperlink to abstract. At the end of the study after all search tasks are completed for both our system and PubMed, experts are given a questionnaire of 21 7-point Likert-scale questions to judge system utility, interface, and search quality. The first 16 questions are taken from a Post Study System Usability Questionnaire (PSSUQ; \citealp{lewis2002psychometric}) widely used in system quality research. The last 5 questions are designed by the authors to evaluate search quality such as overall result relevance and ranking (for the full question list, see App.~\ref{appendix:kb_util}). 
% , a 16-item standardized questionnaire widely used in system quality and user satisfaction research. Each item is a We augment this set of questions with 5 additional questions focused on information retrieval, such as impression with overall relevance and ranking of results. 
Each question is asked twice, once for PubMed and once for our system, leading to 21$\times$2$\times$5 = 210 responses. 

\vspace{.1cm} \noindent {\bf Search queries} We provide experts with seven search queries that were created by an expert medical researcher, relating to causal links (e.g., between COVID-19 and cardiac arrhythmias) and functions (e.g., Ivermectin as a treatment). See full set of queries in App.~\ref{appendix:human_evaluation_guidelines}.
%We provide the experts a short document explaining our interface (available in the supplementary material) =

% This is a challenging task, evaluating both \emph{retrieval} of relations that are semantically similar to the query, and also accurate \emph{extraction} of relations from sentences. 

\vspace{.1cm} \noindent {\bf Results} Fig.~\ref{fig:kb_eval} (right) shows the average Likert scores (normalized to  [0\%,100\%]) across all questions and users for \kb and PubMed. 
% \hanna{maybe no need to normalize; just report the average from 0-7} % how experts   the results of  the results of . We normalize the Likert scale responses to [0\%,100\%] and compute the average across questions and users.
The results show that the medical experts strongly prefer \kb to PubMed (overall average of 91\% vs. 74\%, with non-normalized scores of 6.6 vs. 5.2). On average across the 21 questions, the majority of the five experts assigned our interface a higher score than PubMed, at an average rate of 3.5/5. This rate increases further when considering ties---on average 4.75/5 of the experts assigned our system a score equal or higher than PubMed.
%of finding information focused on search for mechanism/cause-effect relations

Overall, our system significantly outperforms PubMed in this task, with an average gap of roughly 20\% for search and utility-related questions (Wilcoxon signed rank test p-value is significant at $4.77\times10^{-7}$). These results are particularly interesting and indicate the potential of \kb because of the experts' strong familiarity with PubMed and the simple nature of our UI.

Our system searches and retrieves \textit{relations}---only texts explicitly mentioning relations that match the input query. This often more precisely reflects the query than results returned by PubMed, which do not have the additional layer of structured information in \kb. For example, for the query ($E_1$=cardiac arrhythmias, $E_2$=COVID-19), PubMed returns the following title of one paper: \textit{Guidance for cardiac electrophysiology during the \textbf{COVID-19} pandemic [....] Electrocardiography and \textbf{Arrhythmias} Committee}---$E_1$ and $E_2$ are both mentioned, but not within a mechanism relation.

% \hanna{we can remove this last paragraph; not that interesting.}
% In a different example, for the relation (E1=Antibody therapy, E2=COVID-19) a PubMed search specifying that both terms should occur in either title or abstract, returned 5/10 results where at least one term was not in the title or the text snippet underneath it, requiring to read the abstract. A closer look showed that some those 5 papers discussed the queried relation between E1 and E2 across multiple, long sentences that could not be fit in one snippet, hence requiring more effort from the user. In contrast, our system directly retrieves relations, and is able to pull them out of the text and display to users, as illustrated in Figure~\ref{fig:flow} (right).

% Another important factor relates to the structured information retrieved from \kb , such as

\section{Conclusion}
\label{sec:conclusion}
We introduced a unified schema for {\em mechanisms} that generalizes across many types of activities, functions and influences. We constructed and distributed \data, a dataset of papers related to COVID-19 annotated with this schema. We trained an IE model and applied it to COVID-19 literature, constructing \kb, a KB of 1.5M mechanisms. We showcased the utility of \kb in structured search for mechanism relations in COVID-19 literature. In a study with MDs active in the fight against the disease, our system is rated higher than PubMed search for both utility and quality. Our unified view of mechanisms can help generalize and scale the study of COVID-19 and related areas. More broadly, we envision a KB of mechanisms that enables the transfer of ideas across the literature \cite{hope2017accelerating}, such as by finding relationships between mechanisms in SARS-CoV-2 and other viruses, and assists in literature-based discovery \cite{Swanson1996UndiscoveredPK} by finding cross-document causal links.
% We hope our framework can support research on COVID-19, and boost scientific knowledge discovery more broadly.

\section*{Ethical considerations}

Our knowledge-base and search system is primarily intended to be used by biomedical researchers working on COVID-19, and researchers from more general areas across science. Models trained and developed on our dataset are likely to serve researchers working on COVID-19 information extraction, and scientific NLP more broadly. We hope our system will be helpful for accelerating the pace of scientific discovery, in the race against COVID-19 and beyond.

Our knowledge-base can include incorrect information to the extent that scientific papers can have wrong information. Our KB includes metadata on the original paper from which the information was extracted, such as journal/venue and URL. Our KB can also miss information included in some papers.

Our data collection process respected intellectual property, using abstracts from CORD-19 \cite{wang2020cord}, an open collection of COVID-19 papers. Our knowledge-base fully attributes all information to the original papers. All annotators were given extensive background on our objectives, and told their annotations will help build and evaluate a knowledge-base and search engine over COVID-19 research. Graduate-student annotators were payed 25 USD per hour. MD experts helped evaluate the tool on a voluntary basis.

\section*{Acknowledgements}
We like to acknowledge a grant from ONR N00014-18-1-2826. Authors would also like to thank anonymous reviewers, members of AI2, UW-NLP and the H2Lab at The University of Washington for their valuable feedback and comments.

\bibliography{00_main.bib}
\bibliographystyle{acl_natbib}

\appendix

\section{Data Annotation}
\label{appendix:data_anno}
\subsection{ Granular Relations}
\label{appendix:granular}
\begin{figure}[h]
\setlength{\belowcaptionskip}{-12pt} 
    \centering
\includegraphics[width=0.9\linewidth]{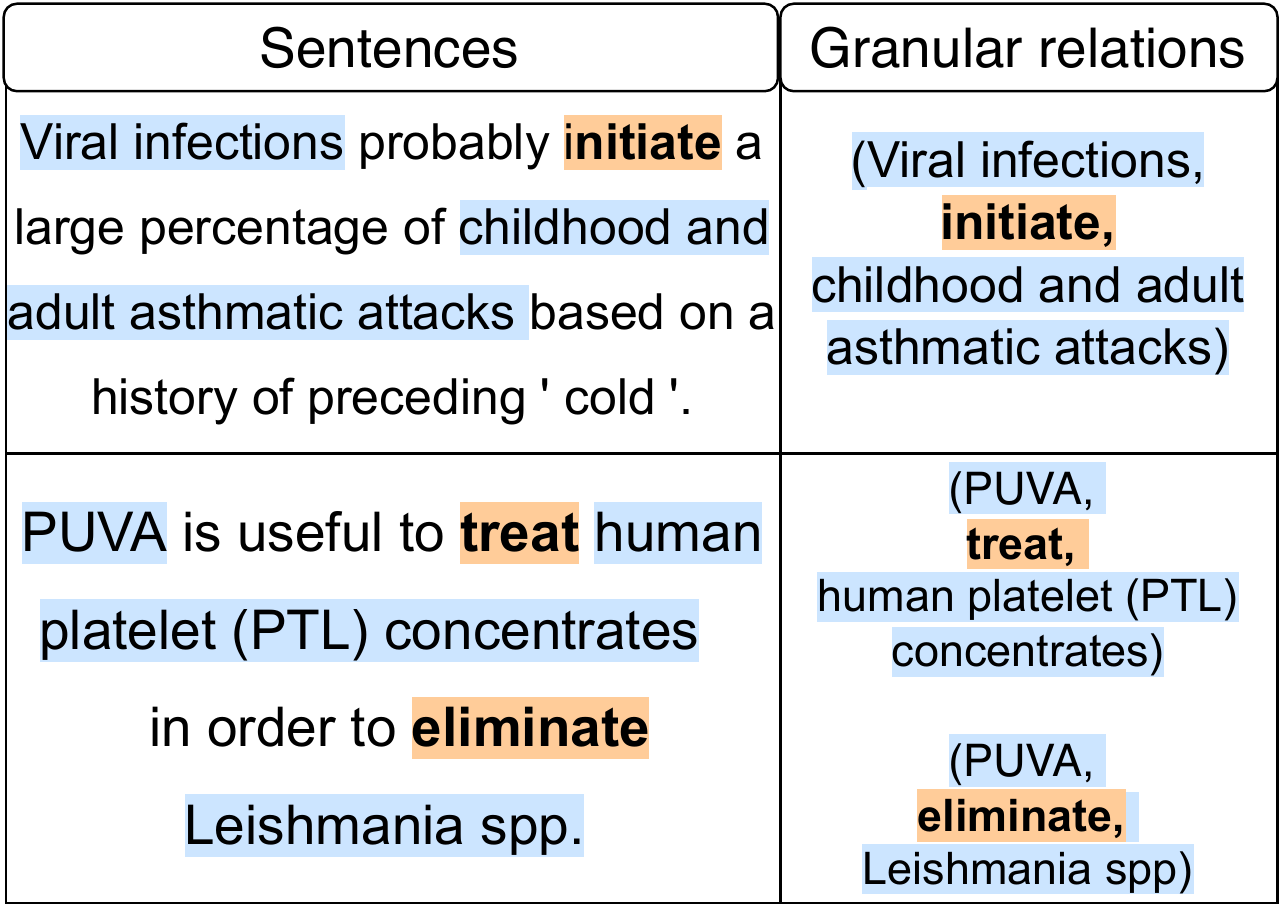}  
\caption{Examples of granular relations.}
    \label{fig:eventfig}
\end{figure}

In addition to the two coarse-grained relation classes, we also experimented with \emph{granular} relations where the {\tt class} represents a specific type of a mechanism relation explicitly mentioned in the text (we constrain the mention a single token for simplicity, e.g., \textit{binds, causes, reduces}; see Fig.~\ref{fig:eventfig} for examples of granular relations). While more granular, these relations are also less general -- as the natural language of scientific papers describing mechanisms often does not conform to this more rigid structure (e.g., long-range and implicit causal relations). We thus focus most of our work on coarse-grained relations. We release our dataset and a model for extraction of granular relations to support future research and applications, in our code repository.

\subsection{Annotation Collection}
\label{appendix:anno_collection}
We utilize the Prodigy \cite{Prodigy:2018} annotation platform which provides the ability to select span boundaries and relations with ease. Each annotator undergoes a training session in which we cover the definitions of spans and relations as well as use of the platform.  See annotation guidelines in our code repository for more details and examples.

\label{appendix:anno_example}

\begin{table*}[h]
    \centering
    \setlength{\belowcaptionskip}{-8pt}
    %\resizebox{\textwidth}{!}{
    \begin{tabular}{|p{7cm}|p{4cm}|p{4cm}|}
        \hline
        \textbf{Context} & \textbf{Annotator 1} & \textbf{Annotator 2} \\
        \hline
         Predicted siRNAs should effectively silence the genes of SARS - CoV-2 during siRNA mediated treatment. & (predicted siRNAs,  silence the genes of SARS - CoV-2,  DIRECT)
        & (siRNAs,  silence the genes of SARS - CoV-2 during siRNA mediated treatment,  DIRECT) \\
        \hline
        Recent reports show that the inhibition of \textbf{NSP4} expression by small interfering RNAs leads to alteration of the production and distribution of other viral proteins and mRNA synthesis ,  suggesting that NSP4 also \textbf{affects} \textbf{virus} \textbf{replication} by unknown mechanisms. & (NSP4,  affects virus replication,  INDIRECT) & (NSP4,  virus replication by unknown mechanisms,  INDIRECT) \\
        \bottomrule
    \end{tabular}
    \caption{Examples of differences between two annotators. The core meaning of the relation is equivalent across both annotators.}
    \label{table:annotation_errors}
\end{table*}

Tab.~\ref{table:annotation_errors} shows examples of differences between annotations, with disagreements in the span boundaries. This reflects the challenging nature of our task with relations between flexible, open entities. 

%In order to address the issue of span boundary variations,   an NLP-expert annotator  Then the output of this set of annotation is shown to a Bio-NLP expert to correct the labels and check underlying bio-related logics. 
% For each correction task,  we show the annotator the relations that already marked by previous person and they have the opportunity to change,  and then accept/reject/ignore.

% \subsection{Human Evaluation of Predictions}
% We show human evaluators relations predicted by our model and the SRL baseline. For each approach,  we sample random relations to display for each document and then shuffle the combined set.  The evalutor is asked either to accept or reject based on if the span boundaries and relation label makes a \emph{meaningful} and \emph{correct} relation.

% \paragraph{\dataE{} top predicate words}
% \label{appendix:granular_trig}
% Figure \ref{fig:trigs} shows the $30$ most frequent predicates in \dataE{} and their frequency in the annotated data.

% \begin{figure}
%     \centering
%     \includegraphics[width=0.8\linewidth]{mechanism/trig_histo.png}
%     \caption{Most frequent predicate words over the \dataE{} dataset.}
%     \label{fig:trigs}
% \end{figure}

\begin{figure*}
    \centering
\setlength{\belowcaptionskip}{-8pt}
\includegraphics[width=0.45\linewidth]{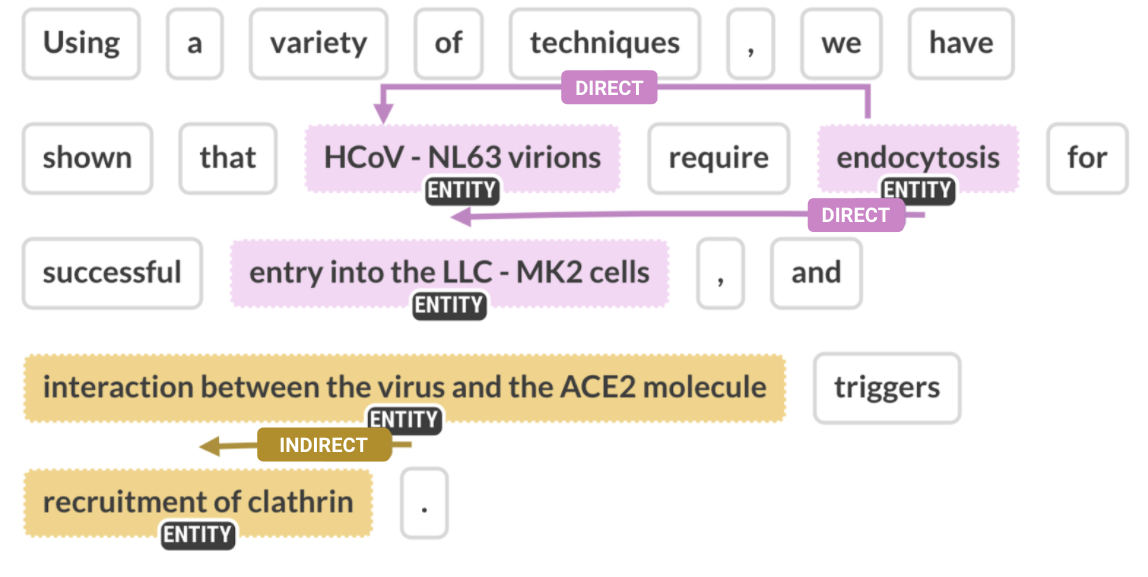}
\includegraphics[width=0.48\linewidth]{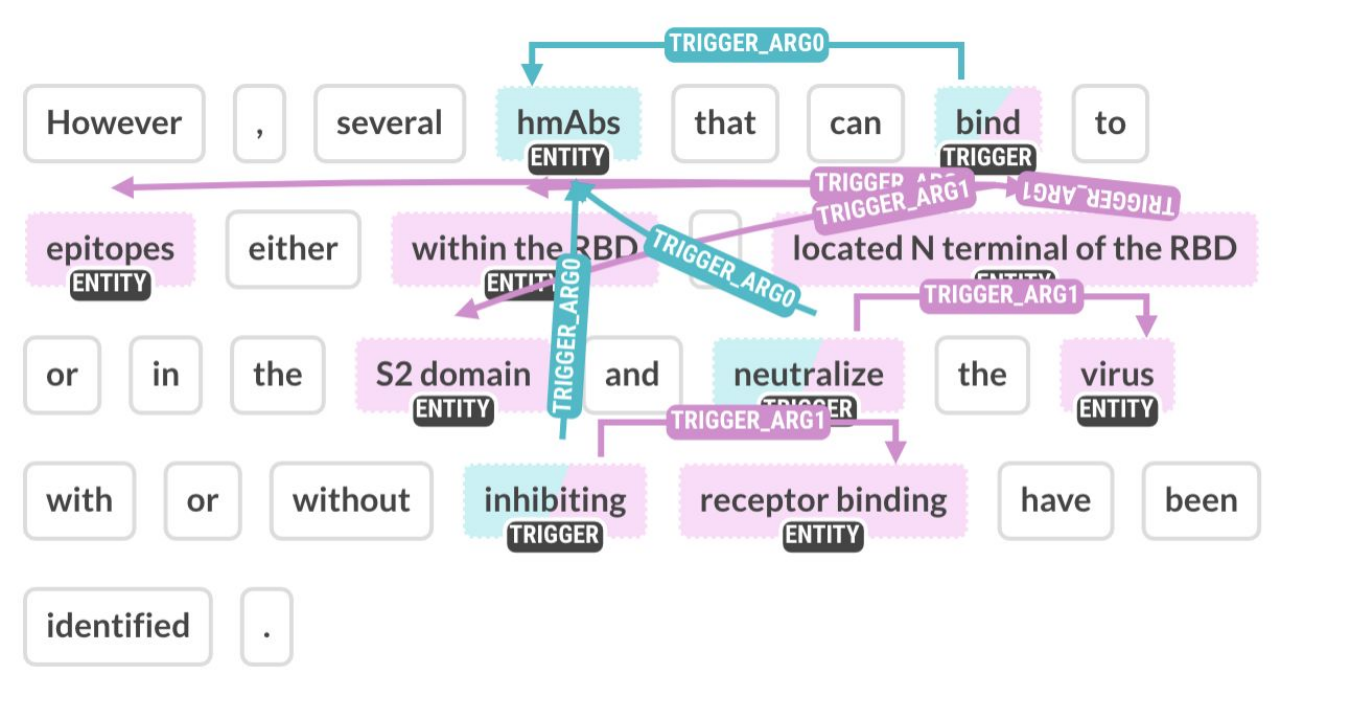} 
\caption{Example of the annotation interface for coarse (left) and granular (right) mechanism relations.}
    \label{fig:annotations}

\end{figure*}

\section{IE Evaluations }

\label{appendix:IE_eval}
\subsection{Automated evaluation metrics}
\label{appendix:eval_metrics}
\vspace{.1cm}\noindent{\bf Entity detection} 
Given a boolean span matching function $\textrm{m}(s_1, s_2) = \mathbbm{1}(s_1$ matches $s_2)$, a predicted entity mention $\hat{e}$ is correctly \emph{identified} if there exists some gold mention $e^*$ in $\mathcal{D}$ such that $\textrm{m}(\hat{e}, e^*) = 1$ (since there is only one entity type, an entity mention is correctly classified as long as its span is correctly identified).

Following common practice in work on Open IE \cite{stanovsky2018supervised}, we report results using a partial-matching similarity function, in this case based on the widely-used Rouge score: $\textrm{m}_{\textrm{rouge}}(s_1, s_2)$ is true if Rouge-L$(s_1, s_2) > 0.5$ \cite{lin2004rouge}.

\vspace{.2cm}\noindent{\bf Relation detection / classification} Given a boolean span matching function, a predicted coarse-grained relation $\hat{r} = (\hat{E}_1, \hat{E}_2, \hat{y})$ is correctly \emph{identified} if there exists some gold relation $r^* = (E_1^*, E_2^*, y^*)$ in $\mathcal{D}$ such that  $\textrm{m}(\hat{E}_1, E_1^*) = 1$ and $\textrm{m}(\hat{E}_2, E_2^*) = 1$. It is properly \emph{classified} if, in addition, $\hat{y} = y^*$.

Relation identification measures the model's ability to identify mechanisms of any type - direct or indirect - while relation classification aims to discriminate between direct and indirect types of mechanism mentions in the text. 

\subsection{Baselines}
\label{appendix:baseline}

\paragraph{SemRep} The SemRep dataset \cite{kilicoglu2011constructing}, consisting of 500 sentences from MEDLINE abstracts and annotated for semantic predication. Concepts and relations in this dataset relate to clinical medicine from the UMLS biomedical ontology \cite{bodenreider2004unified}, with entities such as drugs and diseases. Some of the relations correspond to mechanisms (such as X TREATS Y or X CAUSES Y); By the lead of domain experts, we map these existing relations to our mechanism classes and use them to train DyGIE. Other relations are even broader, such as PART-OF or IS-A -- we do not attempt to capture these categories as they often do not reflect a functional relation.

\paragraph{Scierc} SciERC dataset \cite{luan2018multi}, consisting of 500 abstracts from computer science papers that are annotated for a set of relations, including for USED-FOR relations between methods and tasks. We naturally map this relation to our {\tt DIRECT} label and discard other relation types, and use this dataset to train DYGIE.

\paragraph{SRL} Finally we also use a recent BERT-based SRL model \cite{shi2019simple}. We select relations of the form (\texttt{Arg0, verb, Arg1}), and evaluate using our partial metrics applied to \texttt{Arg0} and \texttt{Arg1} respectively. 

%Finally we also try filtering predicates down to a list of 80 biomedical verbs that are publicly available from a biomedical proposition bank named BioPro \cite{chou2006semi}, map each predicate verb to either \texttt{DIRECT MECHANISM} or \texttt{INDIRECT MECHANISM}, and use this mapping as SRL's predictions (SRL-Mechanism in \ref{fig:IEresults}).

\subsection{Hyperparameter Search}
\label{appendix:hyperparam}
We perform hyperparameter search over these sets of parameters:
\begin{itemize}
    \item {\bf Dropout} is randomly selected from intervals $[0, 0.5]$.
    \item {\bf Learning rate} is randomly selected between $[1e-5, 1e-2]$
    \item {\bf Hidden Size} is randomly selected from interval $[64, 512]$
\end{itemize}
Hyperparameter search is implemented using grid search with the Allentune library \cite{showyourwork}. For each experiment we set the search space to be among $30$ total samples in hyperparameter space. We select the best-performing parameters using the development set.

\subsection{Best Performing Model over \data{}}
\label{appendix:cofie_model}
We use the DYGIE package~\cite{wadden2019entity} to train models for entity and relation extraction over \data{} and we utilize SciBERT \cite{beltagy-etal-2019-scibert} for our text embeddings and finetune upon that, with learning rate for finetuning set to $5e-5$ with weight decay of $0.01$. The training was run for $100$ epochs with the  $slanted\_triangular$ \cite{howard2018universal} learning rate scheduler. We used the AdamW \cite{Loshchilov2017FixingWD} optimization algorithm. In our objective function we assign equal weights to relation and span loss terms. The maximum allowed length of spans is $12$.

The hyperparameters achieving best performance over our development search are $0.43$, $7e-4$ and $215$ for dropout, learning rate and hidden size respectively. All other parameters are kept to default values (available in our code repository).

\begin{figure}[t]
         \includegraphics[width=\linewidth]{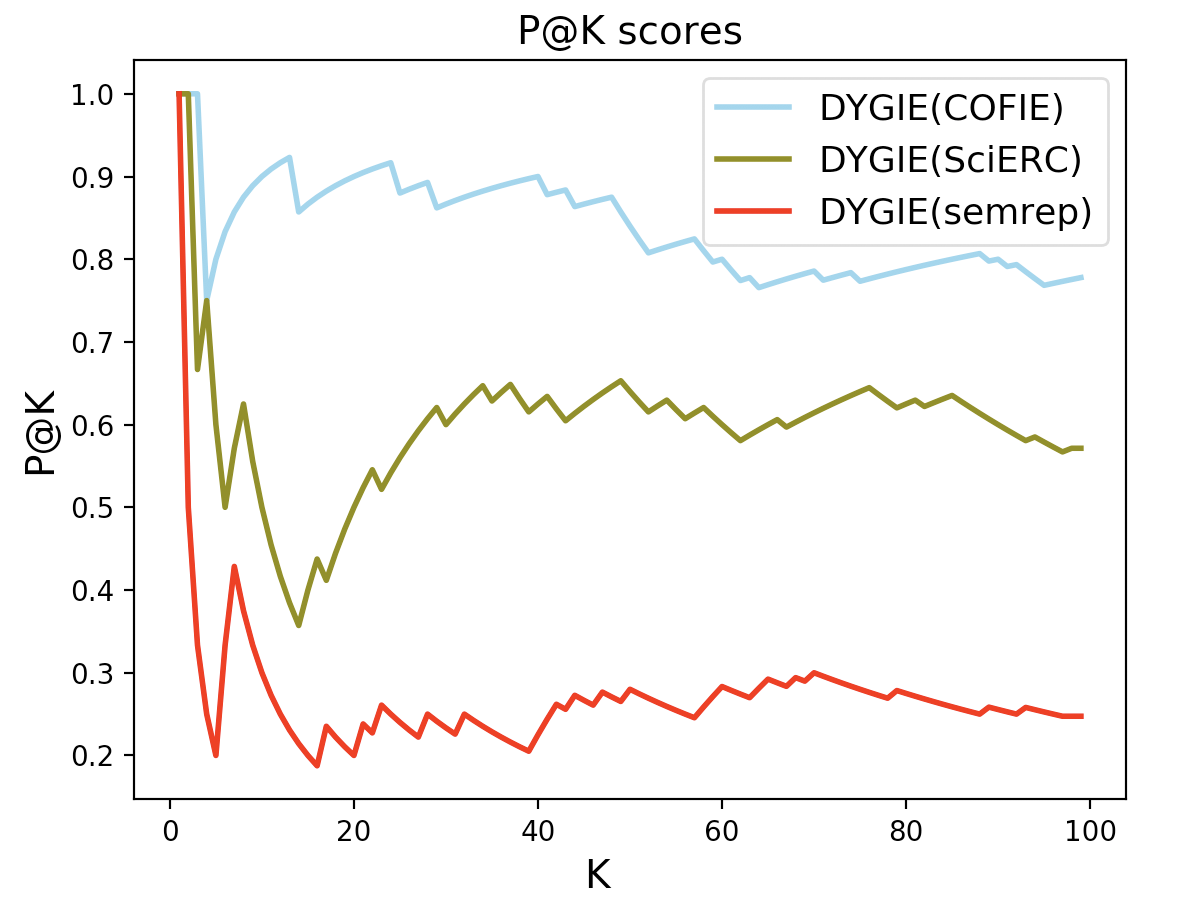}
         \caption{Precision@K  of  our  model  compared  with pre-trained  SciERC and SemRep baselines.  P@K  for  our  model is  high in absolute numbers.}
         \label{fig:p_at_K}
\end{figure}

Tab.~\ref{table:IEresultsie} compares the performance of our best model with the baselines introduced in Sec.~\ref{appendix:baseline}. Fig.~\ref{fig:p_at_K} shows Precision@K results, with our model reaching high absolute numbers.
\begin{table}
\setlength{\belowcaptionskip}{-6pt} 
\centering
\begin{small}
\begin{tabular}{l|rrr}
\toprule
Model &\multicolumn{1}{c}{\textbf{RC}} & \multicolumn{1}{c}{\textbf{RD}} & \multicolumn{1}{c}{\textbf{ED}} \\
\toprule
% Model & Substr &  Rouge &    Exact &      Substr &  Rouge &    Exact & Substr &  Rouge &    Exact\\
% \midrule
OpenIE    &   - &  15.5 & 25.6 \\\
SRL &    -  &  24.5 &  27.7  \\

\midrule
DYGIE(SemRep)   &   6.8 &   8.3 &  32.5   \\
DYGIE(SciERC) &  18.6 &   20.4 &   39.2  \\
\midrule
DYGIE(\data)    &  \textbf{42.8} & \textbf{45.6} &  \textbf{50.2} \\
\end{tabular}
\end{small}
\caption{F1 scores. Relations from SRL and OpenIE do not map directly to {\tt DIRECT MECHANISM} and {\tt INDIRECT MECHANISM} classes, and do not have relation classification scores.}
\label{table:IEresultsie}
\end{table}

\subsection{Granular relation prediction}
\label{appendix:granular_preds}
Granular relations are evaluated in the same fashion as coarse-grained relations, with the additional requirement that the predicted predicate token $\hat{p}$ must match the gold $p^*$.
Our evaluation shows that the model trained to predict granular triples achieves $F1$ score of $44.0$. When predicting relations without trigger labels (i.e., {\tt (s, o)}), the model achieves $F1$ scores of $53.4$. These results are not comparable to those for \data{}, which includes more documents and relations that did not directly conform to the {\tt (s, o, p)} schema.

\subsection{Best Performing Model over \dataE{}}
\label{appendix:granular_model}
Here too we use the DYGIE package \cite{wadden2019entity} with SciBERT \cite{beltagy-etal-2019-scibert}. Due to technical equivalence in the annotation schema of our granular relations and the event extraction task in \citet{wadden2019entity}, we make use of the event extraction functionality of DYGIE. For fine-tuning the embedding weights of SciBERT we used the same learning rate weight as for \data{}, and the best hyperparameters found are $0.30$, $11e-4$ and $372$ for dropout, learning rate and hidden layer size respectively. All other parameters are kept to default values (available in our code repository).

\section{Human evaluation guidelines}\label{appendix:human_evaluation_guidelines}

\subsection{KB Correctness and Informativeness evaluation guideline}

\label{appendix:KB_eval}
% {\bf See task guidelines attached at the bottom of this document} including examples given to annotators. 

\paragraph{Relation quality evaluations over various domains} For the task involving the exploration of viral mechanisms,  we used 10 recent scientific claims taken from \cite{wadden2020fact}. These 10 claims,  and the queries constructed for them,  are as follows:

\begin{itemize}
    \item Remdesevir has exhibited favorable clinical responses when used as a treatment for coronavirus.  X = [Remdesevir], Y = [SARS-CoV-2, coronavirus, COVID-19]

    \item Lopinavir / ritonavir have exhibited favorable clinical responses when used as a treatment for coronavirus.  X = [Lopinavir, Ritonavir], Y = [SARS-CoV-2, coronavirus, COVID-19]

    \item Aerosolized SARS-CoV-2 viral particles can travel further than 6 feet. X = [Air, Aerosols, Droplets, Particles, Distance], Y = [SARS-CoV-2 transmission]

\item Chloroquine has shown antiviral efficacy against SARS-CoV-2 in vitro through interference with the ACE2-receptor mediated endocytosis. X = [Chloroquine], Y = [ACE2-receptor, Endocytosis, interference with the ACE2-receptor mediated endocytosis.]

\item Lymphopenia is associated with severe COVID-19 disease. X = [Lymphopenia], Y = [severe COVID-19 disease, COVID-19]

\item Bilateral ground glass opacities are often seen on chest imaging in COVID-19 patients. X = [Bilateral ground glass opacities], Y = [chest imaging in COVID-19 patients]

\item Cardiac injury is common in critical cases of COVID-19. X = [COVID-19], Y = [Cardiac injury]

\item Cats are carriers of SARS-CoV-2. X = [Cats], Y = [SARS-CoV-2]

\item Diabetes is a common comorbidity seen in COVID-19 patients. X = [Diabetes], Y = [COVID-19]

\item The coronavirus cannot thrive in warmer climates. X = [warmer climates], Y = [coronavirus]

\item SARS-CoV-2 binds ACE2 receptor to gain entry into cells. X = [SARS-CoV-2], Y = [binds ACE2 receptor, binds ACE2 receptor to gain entry into cells]
\end{itemize}

For the AI open-ended search task, we used the following approaches/areas as queries (see guidelines and examples in our code repository): \textit{artificial intelligence, machine learning, statistical models, predictive models, Graph Neural Network model, Convolutional Neural Network model, Recurrent Neural Network model, reinforcement learning, image analysis, text analysis, speech analysis}.

\begin{figure}[t]
         \includegraphics[width=\linewidth]{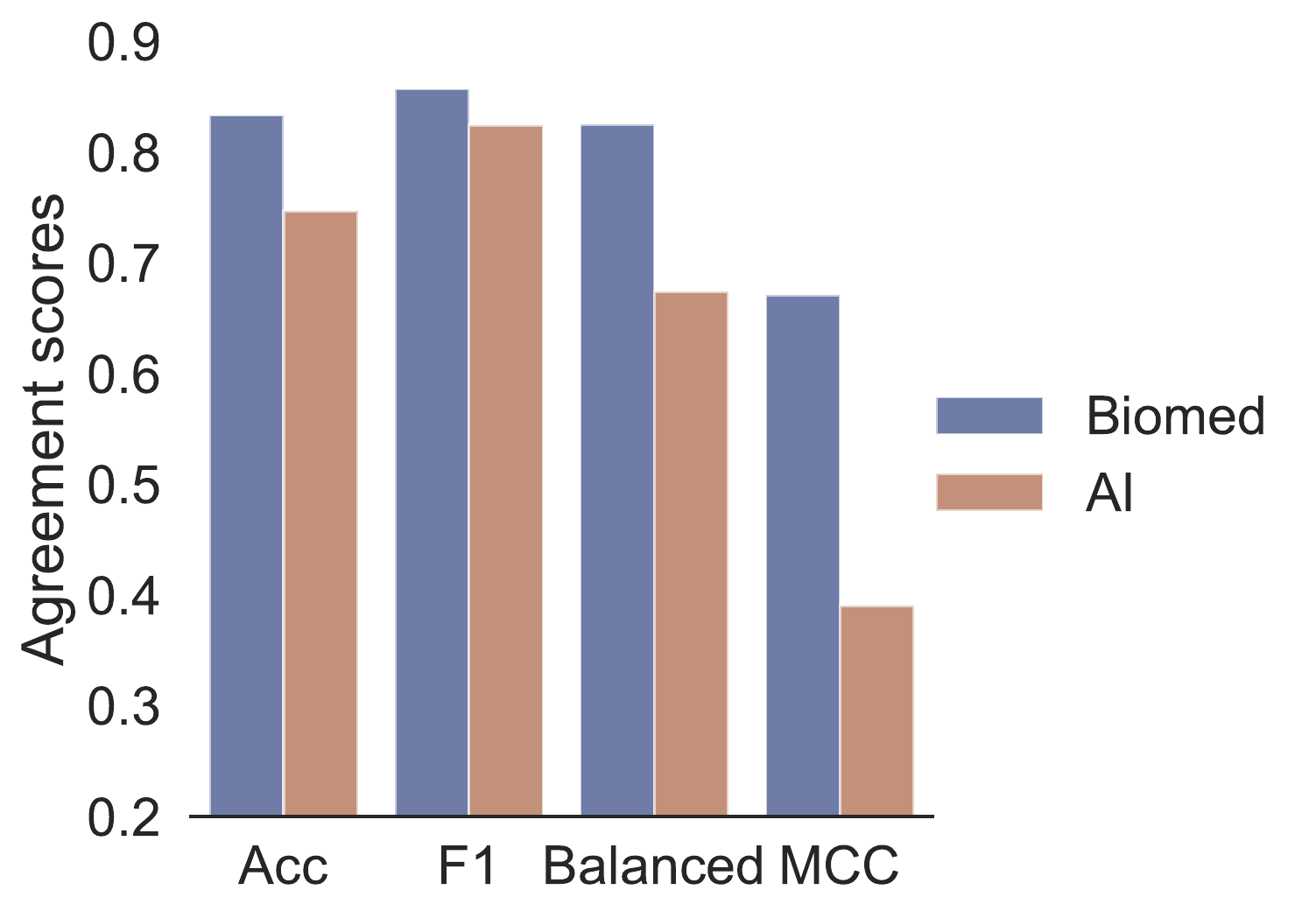}
         \caption{Average pairwise annotator agreement by several metrics. In the AI task human labels were more diverse but with overall high precision / recall.}
         \label{fig:anno_agree}
\end{figure}

For both tasks, we use the following metrics to measure pairwise agreement between annotators (Fig.~\ref{fig:anno_agree}): standard accuracy (proportion of matching rating labels), F1 (taking into account both precision and recall symmetrically), balanced accuracy (with class weights to down-weight the higher proportion of positive ratings), and finally the Matthew Correlation Coefficient (MCC) score, using the corresponding functions in the Scikit-Learn Python library \cite{scikit-learn}.

%  We measure average pairwise annotator agreement with several metrics: accuracy (proportion of matching labels), F1 (taking into account precision and recall symmetrically), balanced accuracy (down-weighting the positive ratings to counter their higher proportion), and the Matthew Correlation Coefficient (MCC) score.

\paragraph{Comparing  KB  quality  to  other  schema}
We sampled the relations predicted by our model and the baseline models introduced in App.~\ref{appendix:baseline}. We randomly selected 20 abstracts from the \data{} test set and show at most two predictions (if available) for each sentence within that abstract. In total 300 relations are extracted. 
Each relation is shown separately to two bio-NLP expert annotators (with annotators blind to the condition), who label each relation with a 0/1 label (1 if the relation is both \emph{correct} and \emph{informative}).

\subsection{KB Utility}
\label{appendix:kb_util}
MDs are instructed to search with our interface and with PubMed search, with the following 7 topics:
\begin{itemize}
    \item Query 1: Cardiac arrhythmias caused by COVID 19
    \item Query 2: Hydroxychloroquine and its effect on COVID 19
    \item Query 3: Ivermectin and its role in management of COVID 19
    \item Query 4: Pulmonary embolism effect on complications related to COVID 19
    \item Query 5: Liver disease and COVID 19
    \item Query 6 : Inflammatory bowel disease and COVID -19
    \item Query 7 : Antibody therapy and its uses/effects on COVID-19
\end{itemize}

\begin{figure}
    \centering
    \includegraphics[width=\linewidth]{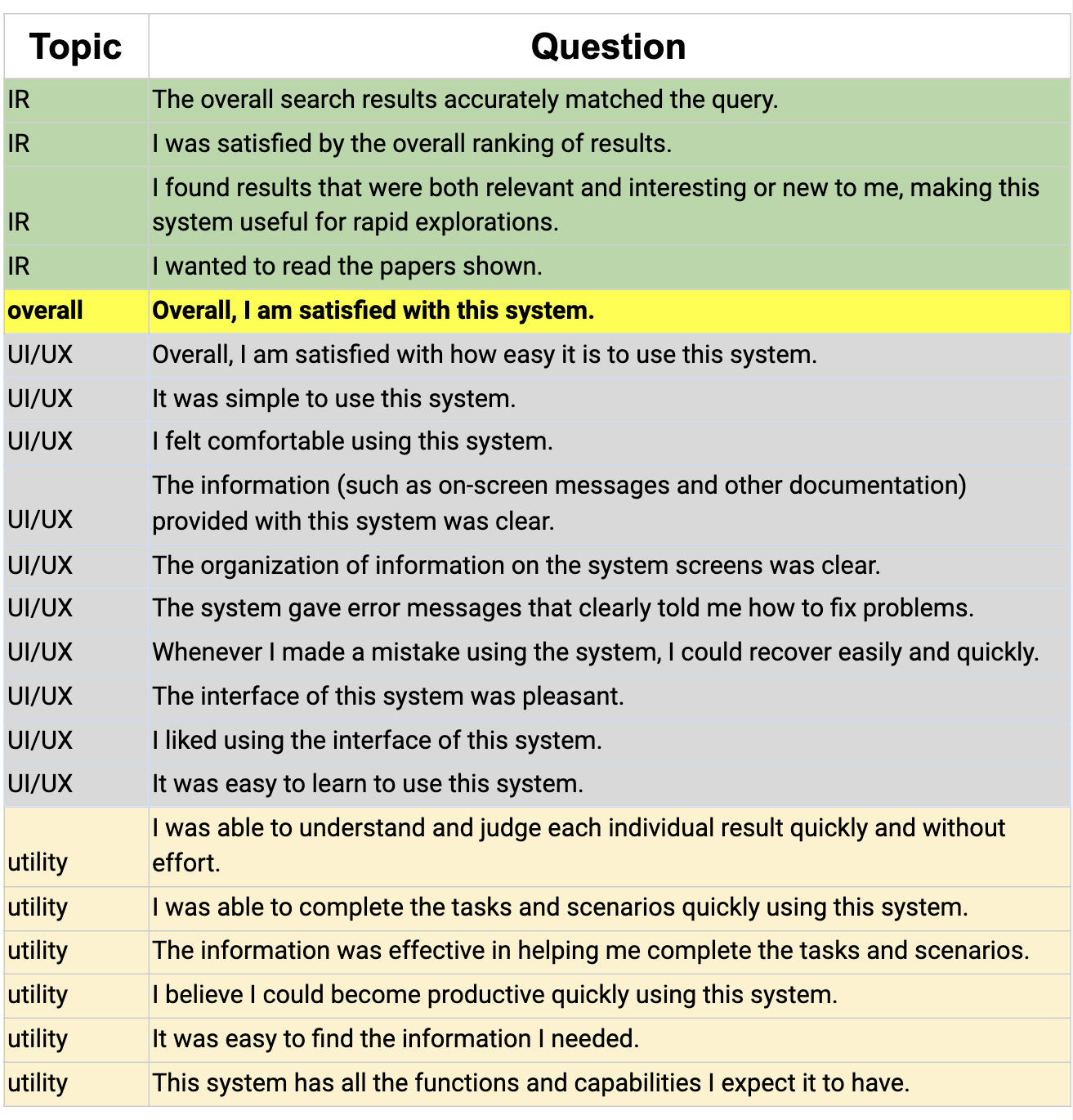}
    \caption{List of post-study questions given to MDs.}
    \label{fig:kb_survey}
\end{figure}

The full list of the post-study evaluation questions given to MDs is shown in Fig.~\ref{fig:kb_survey}.

\end{document}